\pdfoutput=1

\documentclass[11pt]{article}

\usepackage[]{ACL2023}

\usepackage{times}
\usepackage{latexsym}
\usepackage{booktabs}
\usepackage[T1]{fontenc}
\usepackage{enumitem}

\usepackage[utf8]{inputenc}

\usepackage{microtype}
\usepackage{comment}
\usepackage{graphicx}
\usepackage{clrscode}

\newcommand\tow{\textit{The Outer Worlds}}
\newcommand\ow{\textit{Outer Worlds}}
\newcommand{\name}{\textsc{Knudge}}
\newcommand{\knudge}{\name}

\newsavebox\myv

\newcommand\blfootnote[1]{%
  \begingroup
  \renewcommand\thefootnote{}\footnote{#1}%
  \addtocounter{footnote}{-1}%
  \endgroup
}

\usepackage{array}
\makeatletter
\newcolumntype{V}[1]{>{\topsep=0pt\@minipagetrue}p{#1}<{\vspace{-\baselineskip}}}
\makeatother
\newcolumntype{L}[1]{>{\raggedright\let\newline\\\arraybackslash\hspace{0pt}}m{#1}}

\usepackage{listings}
\usepackage{longtable}

\newcommand{\tr}[1]{\textcolor{purple}{#1}}
\newcommand{\tb}[1]{\textcolor{blue}{#1}}
\newcommand{\tor}[1]{\textcolor{orange}{#1}}
\title{Ontologically Faithful Generation of Non-Player Character Dialogues}

\author{Nathaniel Weir$^\spadesuit$\thanks{\ \ Work done as an intern at Microsoft Semantic Machines.}  \quad Ryan Thomas$^\dag$ \quad Randolph d'Amore$^{\ddag}$ \quad Kellie Hill$^{\dag}$   \\ 
\textbf{Benjamin Van Durme$^{\spadesuit\dag}$}\quad \textbf{Harsh Jhamtani$^{\dag}$}  \\
$^\spadesuit$Johns Hopkins University \quad $^\dag$Microsoft Semantic Machines
\quad $^\ddag$Microsoft Gaming\\
\texttt{nweir@jhu.edu} \quad \texttt{hjhamtani@microsoft.com}
}

\begin{document}
\maketitle
\begin{abstract}

We introduce a language generation task grounded in a popular video game environment. \name{} (\textbf{KN}owledge Constrained \textbf{U}ser-NPC \textbf{D}ialogue \textbf{GE}neration) requires models to produce trees of dialogue between video game characters that accurately reflect quest and entity specifications stated in natural language.
\name{} is constructed from side quest dialogues drawn directly from game data of Obsidian Entertainment’s \textit{The Outer Worlds}, leading to real-world complexities in generation: 
(1) dialogues are branching trees as opposed to linear chains of utterances; (2) utterances must remain faithful to the game lore-- character personas, backstories, and entity relationships; and (3) a dialogue must accurately reveal new quest details to the human player. 
We report results for a set of neural generation models using supervised and in-context learning techniques; we find competent performance but  room for future work addressing the challenges of creating realistic, game-quality dialogues.

\end{abstract}

\section{Introduction}
\begin{figure}[t]
    \centering
    \includegraphics[width=\columnwidth]{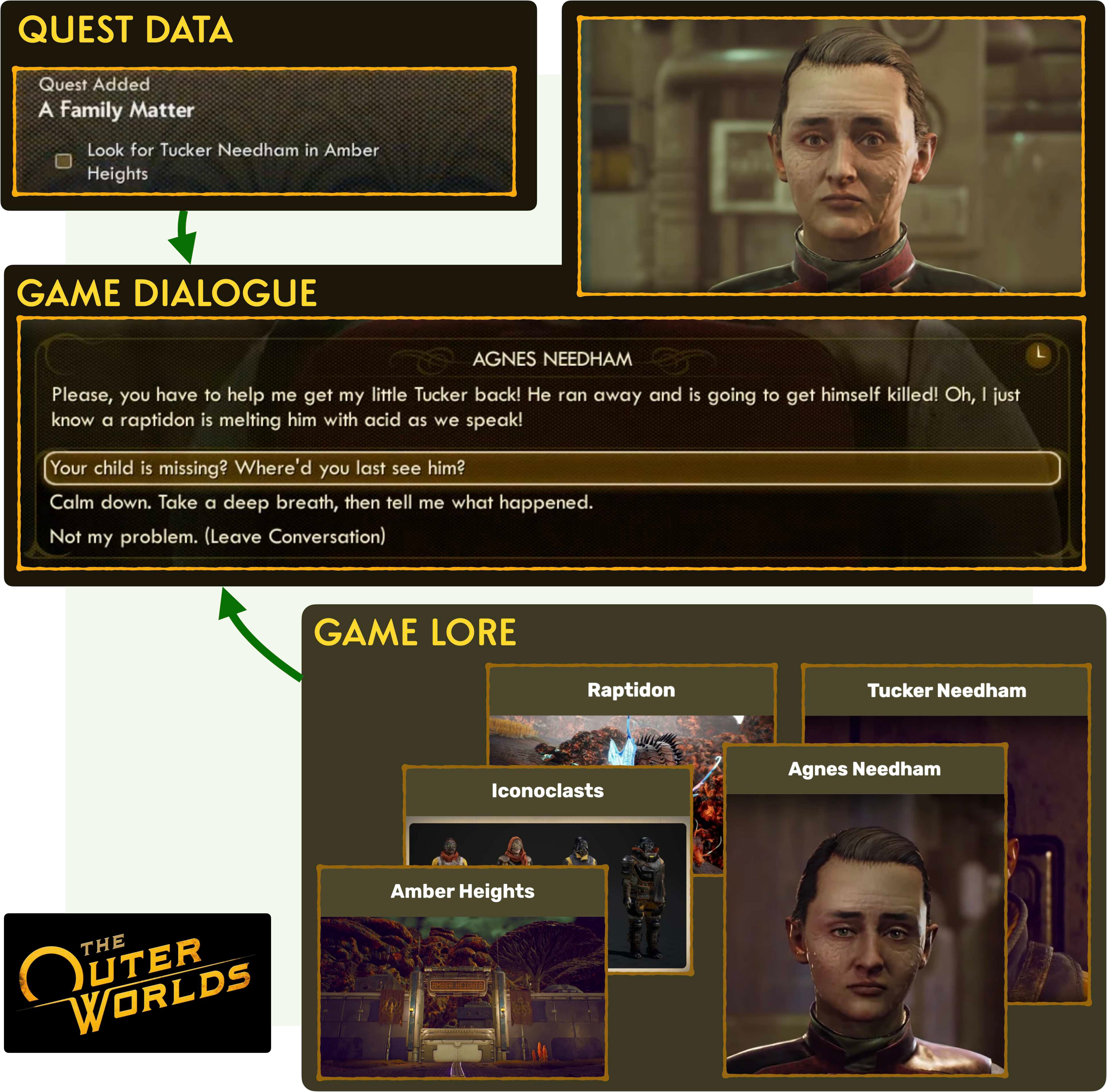}
    \caption{An example non-player character (NPC) dialogue from \tow{}~\cite{game:theouterworlds}. NPCs  must speak faithfully to a granular ontology of \textbf{quest specifications} and \textbf{game lore}.}
    \label{fig:teaser}
\end{figure}
\blfootnote{** \name{} will be publicly available
 upon publication.
} 
Player interactions with non-player characters (NPCs) in role-playing games (RPGs) often serve %
to flesh out backstories %
while allowing the player to progress through engaging quest storylines~\cite{onuczko2007demonstration}. 
A key challenge in authoring these NPC dialogues is maintaining enjoyable as well as \emph{coherent} narratives: utterances must faithfully reflect quest structure and game lore-- characters, histories, and entity relationships. 

\autoref{fig:teaser} depicts a dialogue turn taken from \tow{}~\cite{game:theouterworlds},\footnote{\url{https://en.wikipedia.org/wiki/The_Outer_Worlds}} an action RPG renowned for its narrative and dialogue writing. This turn demonstrates how a dialogue relies on the descriptions of entities in the game world while also revealing relevant quest information. The NPC's utterance not only begins and provides backstory about a new side quest,\footnote{RPG \emph{side quests} are goals provided to a player in order to enrich gameplay, while not being part of the central storyline.} but also interacts according to a well-formed persona (a worried parent) and references an adversary (raptidons) that the player will face later in the quest.

NPC interactions often take the form of complex utterance trees. %
Creating these branching structures according to the many specifications of dialogue writing can be time-consuming for a game designer \cite{Caropreso2012TemplateAE}, and costs millions of dollars (see \S\ref{app:cost} for discussion).
This motivates the pursuit of \textit{automatically generating} dialogue trees.

To the best of our knowledge, there does not exist a public dataset that meets our desired criteria: \textbf{a set of real game-quality NPC dialogues paired with granular quest and biographical specifications consistent with a well-formed game ontology.}
Recent work towards lore-conditioned dialogue~\cite{DBLP:conf/emnlp/UrbanekFKJHDRKS19,DBLP:conf/fdg/StegerenM21}, 
story generation~\cite{DBLP:conf/emnlp/AkouryWWHPI20, chen-gimpel-2022-leveraging},
and knowledge conditioning for task-oriented dialogue agents~\cite{DBLP:conf/emnlp/ChoiHIYYCLZ18,DBLP:conf/emnlp/MazareHRB18, DBLP:conf/emnlp/FengWGPJL20} 
do not address complex dialogue trees and the interweaving narratives found in deployed RPGs.

We make the following contributions via this paper:
firstly, we introduce \name{}: \textbf{KN}owledge constrained \textbf{U}ser-NPC \textbf{D}ialogue \textbf{GE}neration, a set of dialogue trees (in English) derived from an existing video game and paired with granular ontological constraints.
We extract the trees directly from the game data of \textit{The Outer Worlds}.
This game's side quests share overlapping characters and locations, making it an appealing  study in the development of automatic dialogue generation tools. %
For each side quest, we enumerate the relevant dialogues and annotate them at multiple levels of specificity (down to the individual node) with quest- and lore-related natural language (NL) support facts pulled from fan-written wikis. 
\name{} contains 159 dialogues from 45 side quests that all take place in \textit{The Outer Worlds}. It contains 4.7K utterances and an average of 1.3K input constraint tokens per dialogue. The complexity of the dialogues, the annotation level, and the amount of specifications are greater than any comparable dataset.

Secondly, we introduce the challenging task of knowledge-constrained NPC dialogue generation.
Given a set of ontological specifications, we want to generate fluent dialogue trees that reveal the quest objectives while staying faithful to the specified context and the game lore. 
These specifications make up hundreds of statements per dialogue, posing a difficult challenge for modern generation models.
Additionally,
we introduce a class of model, termed DialogueWriters, that 
leverage neural language models (LMs) such as GPT-3 \cite{DBLP:conf/nips/BrownMRSKDNSSAA20} to generate an utterance conditioned on an existing partial dialogue tree and ontology passages. 
To \textit{encourage} the use of game lore to produce interesting and engaging dialogue, we experiment with adding mechanisms to sub-select relevant facts before generating each utterance.

Finally, we prescribe protocols for the evaluation of models for the NPC dialogue generation task.
We test for models' capacity to reflect game ontology constraints in addition to generating fluent and coherent dialogue. 
We conduct automatic and human evaluations of utterances and multi-turn trees generated from specifications for quests from the game, as well as for a pair of never-before-seen quests written by a professional game designer to target real quest writing.
Our experiments reveal further room for improvement in aspects such as joint reasoning over multiple facts from ontology, 
and coverage of all quest objectives. 
We hope that \name{} will facilitate further progress on faithful game dialogue generation, towards tools that can assist developers with their craft.

\begin{figure}[t]
    \centering
    \footnotesize
    \includegraphics[width=\columnwidth]{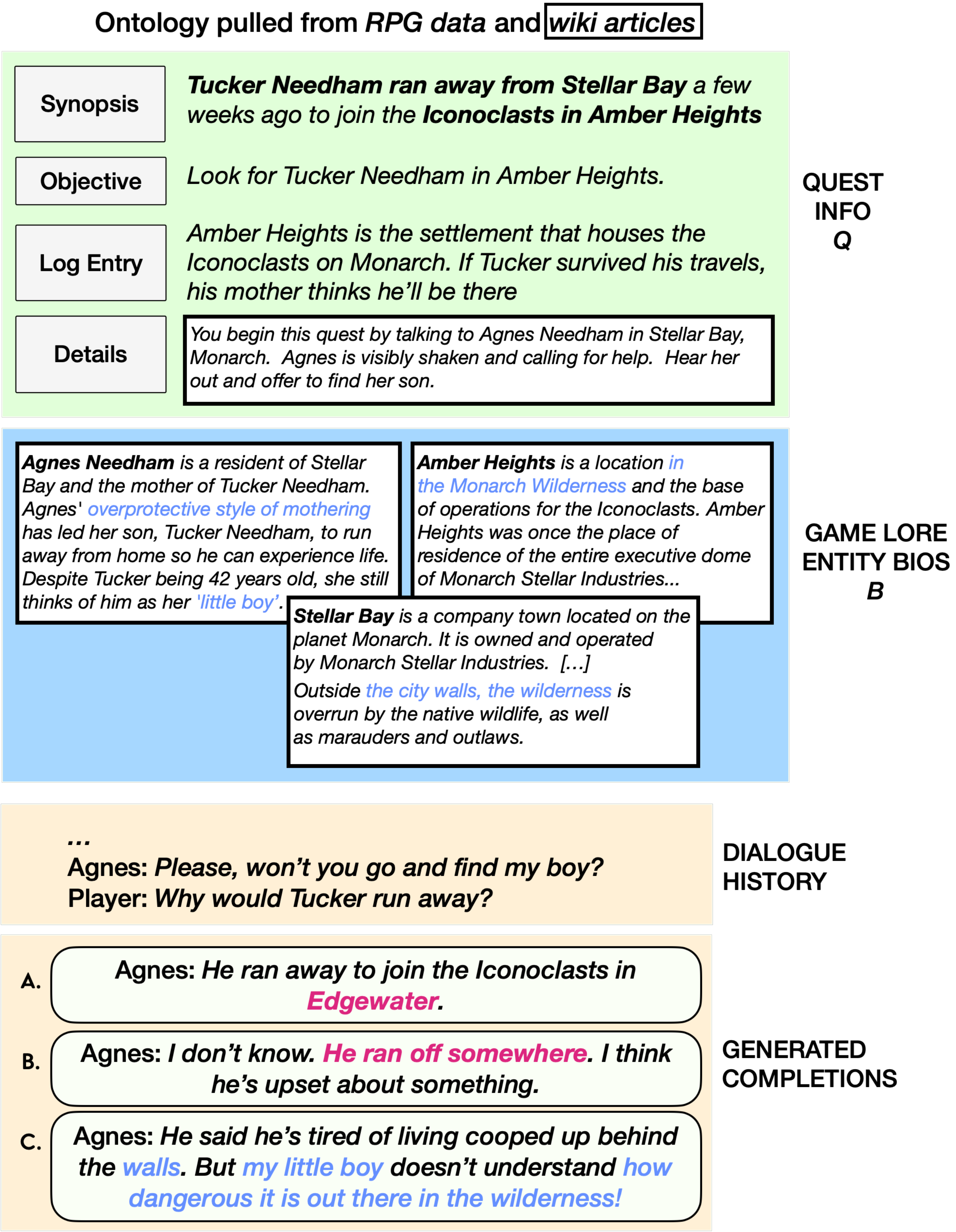}
    \caption{ Overview of the proposed task. Quest information and biographical passages about the game lore serve as constraints on generated dialogue candidates. Completion \textbf{A} is inconsistent with the lore and \textbf{B} is uninformative; \textbf{C} is most desirable, as it provides new information to the player about quest objectives and reflects information about relevant entities.}
    \label{fig:overview}
\end{figure}
\section{Task Definition}
\label{sec:task}
We define the task of knowledge-constrained NPC dialogue generation as the mapping from a set of quest constraint statements $Q$, a set of biographical constraint statements $B$, and a list of participant names $P$  to a dialogue tree $D$.
We consider two task scenarios: \textbf{next utterance prediction} from a partial (gold) tree, and \textbf{full dialogue generation} of trees with multiple candidates at each turn.

As depicted in \autoref{fig:overview} (upper),
$Q$ comprises statements $[q_1, \dots, q_m]$ about currently active objectives upon entering the dialogue, about what should occur during it (e.g., pieces of information the NPC should mention), and about new active objectives upon leaving it. 
$B$ comprises background statements $[b_1, \dots, b_n]$ about game entities that the dialogue must reflect (\autoref{fig:overview}, middle). 
The participant list $P$ contains the player character and one or more NPCs who have corresponding facts in $B$. 
We design these inputs to reflect the kinds of specifications that a developer would provide to the generator during their quest writing process.

Dialogue tree $D$ is a directed graph $\langle N, E \rangle$; each utterance node $n \in N$ has a speaker $s \in P$. 
Branches occur due to the multiple dialogue options at player turns (see \autoref{fig:annotation}, right).
Dialogue trees have one start node, but can have multiple exit nodes and can contain cycles.\footnote{Although they often contain cycles these structures are still colloquially referred to as ``trees''.}\footnote{In \tow{}, utterance nodes are spoken once and are not repeated upon second traversal, unless they are repeat choices by the player at a given decision point. 
}

\section{Data}
\name{} comprises NPCs dialogue trees from all 45 side quests in the \ow{} base game. 
Our dataset construction procedure entails  gathering information about each quest ($Q$) and the \ow{} entities that appear or are referenced during the associated dialogues ($B$) (\S\ref{sec:construction}), and then extracting trees ($D$) from the game data semi-automatically (\S\ref{sec:tree-extraction}). We then analyze the resulting data 
and compare to related datasets (\S \ref{sec:data-comparison}).

\subsection{Game Ontology}
\label{sec:construction}
We acquired dialogue files from the \ow{} creators along with permission to release them publicly; we use quest data and game lore from fan wikis, where a quest's page lists the in-game objectives and journal logs (though the framework allows for using data from official channels instead of fan-made). 
\S\ref{app:dataset} contains further source details.

\paragraph{Quest Information}
A quest in \tow{} appears in the player's journal with a high-level \textbf{synopsis} and a sequence of \textbf{objectives}, each of which contains \textbf{game logs} providing additional details. 
Active objectives are often completed, and new ones are introduced, during an NPC dialogue.
A detailed quest anatomy can be found in appendix \autoref{fig:quest_anatomy}.

We associate with each objective a \textbf{walkthrough passage} which includes details on the topics, player utterance options, and quest information that the NPC needs to say by the dialogue's end.
Thus, a dialogue's quest passage set $Q$ contains:
\begin{enumerate}[itemsep=-1ex]
    \vspace{-1mm}
    \item The \textbf{synopsis} (1-2 sentences)
    \item The \textbf{in objective(s)} active {when entering the dialogue} (1 sentence), the associated \textbf{game log} (1-2) and \textbf{walkthrough passage} (3-10).
    \item The \textbf{out objective(s)} active {upon leaving} the dialogue, and the associated \textbf{game log}.\footnote{We do {not} associate its walkthrough passage, since the NPC should only be expected to convey new objective information that the player will actually see in game.}
\end{enumerate}
Examples of $Q$ can be found in \S\ref{app:quests}. 

\paragraph{Biographical Information}
We associate with each quest, and in turn each dialogue, a set $B$ of \textbf{biographical passages} %
about entities appearing or referenced during the quest.
We extract passages from entities' fan wiki pages. 
While some are only a few sentences, others can be much longer (up to 27), posing a challenge to generation models; often only part of a long biography might be relevant to a given quest.
Examples are shown in \S\ref{app:biographies}.
\subsection{Dialogue Trees}
\label{sec:tree-extraction}
Dialogue trees in \tow{} are complex directed graphs, containing many conditional utterance options depending on the state of the game-- 
e.g. whether the player is of high enough level at some skill to pass a ``check.''\footnote{E.g. if the Player has above 55 points of the \textbf{Persuade} conversational skill, they can convince Tucker Needham to return to his mother in the quest from \autoref{fig:teaser}.} 
To extract a more tractable, quest-related subgraph, we 1) identified the nodes that start and end the interaction using online playthrough videos as reference, and then 2) traversed the graph from the start node, following only edges without special state-related conditions. Special edges were then added manually depending on whether conditions are relevant to the quest.
Example trees can be found in \S\ref{app:dialogues}.
\begin{figure}[t!]
    \centering
    \includegraphics[width=.9\columnwidth]{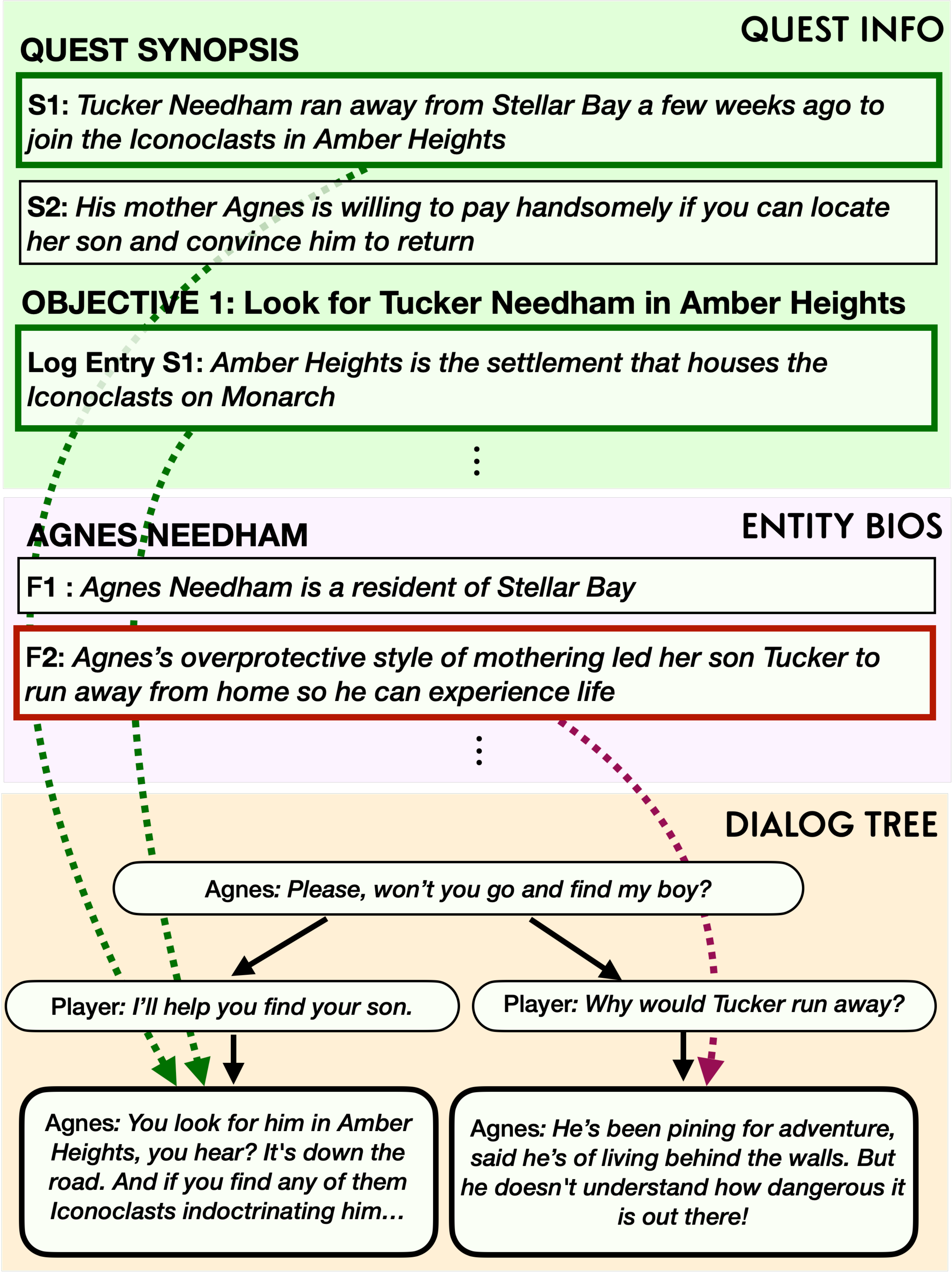}
    \caption{Overview of dialogue node annotation with support facts from quest and biography passages.}
    \label{fig:annotation}
\end{figure}
\paragraph{Annotating Utterance Nodes with Support Facts}
We coordinated with professional data annotators to annotate the nodes of the dialogue trees with support facts from the associated quest and lore constraints. All annotations were done by native English speakers, who were provided with a set of instructions shown in \S\ref{app:human-eval}. We follow a heuristic based on counterfactuals: had a given fact \textit{not} been included in the constraining knowledge, would the utterance be much less likely to occur?
An example of this procedure is depicted in \autoref{fig:annotation}.

We compute an average exact match (EM) score of $0.52$ and an average Jaccard overlap score of $0.62$. These scores represent a high agreement on a subset selection problem where the total set size is often of the order of hundreds of elements. %
\subsection{Dataset Analysis}
\label{sec:data-analysis}
\autoref{tab:statistics} describes overall statistics for the extracted dialogue trees and annotations. We note the high frequency of support fact annotations per utterance; 47\% of all nodes (57\% of NPC nodes) in the dataset are annotated with at least one fact. Amongst NPC nodes, for which support knowledge is more common, there is an average of 1.0 facts per utterance. 
Dialogues have on average 9.9 quest facts and 73.3 biographical facts that generation models must consider when producing utterances.

\begin{table}[t!]
    \centering
    \footnotesize
    \begingroup
    \setlength{\tabcolsep}{2.5pt} %
    \begin{tabular}{lr|lr}
    \toprule 
     Quests    &  45 & Dialogue trees & 159\\
     Entities  & 168 & Facts per dialogue & 83.3\\
     \qquad {{Characters}} & 81 & \qquad Entity facts & 73.3\\
     \qquad {{Locations}} & 40 & \qquad Objective facts & 9.9\\
     \qquad {{Groups}} & 21 & Fact tokens per dialogue & 1321.8 \\
     \qquad {Items} & 18 & Utterances per dialogue & 29\\
     \qquad {{Creatures}} & 7 & \qquad {NPC utterances} & 16.7  \\
     Facts per entity & 7.4 & \qquad {Player utterances} & 12.3\\
     Entities per quest & 8.0 &  Utt. tokens per dialogue & 406.6\\ 
     Quests per entity & 2.0 & Facts per NPC utterance & 1.0\\ 
     Facts per objective & 7.6 & \qquad {{Entity Facts}} & 0.57\\
     Objectives per quest & 4.5 & \qquad {{Objective Facts}} & 0.42\\

     \bottomrule
    \end{tabular}
    \endgroup
    \caption{\name{} Dataset Statistics}
    \label{tab:statistics}
\end{table}

\begin{table*}[t!]
    \scriptsize
    \centering
    \begingroup

    \setlength{\tabcolsep}{2.4pt}
    \renewcommand{\arraystretch}{1.35}
    \begin{tabular}{L{2.8cm}L{1.4cm}L{1.3cm}L{1.9cm}L{.8cm}L{1.2cm}L{3.55cm}L{1.3cm}} 
    \toprule
    \textbf{Dataset} & \textbf{Writer} & \textbf{Source} & \textbf{Structure} & \textbf{Tokens \newline / Item } & \textbf{Constraint \newline Toks / Item }  & \textbf{\tor{Narrative} and \tb{Bio} Constraints} & \textbf{Annotation Level} \\ \midrule
    \multicolumn{2}{l}{\textbf{Story Generation Datasets}} \\ \midrule 
\textsc{storium}\newline \cite{DBLP:conf/emnlp/AkouryWWHPI20} & Crowd & Online story writing game & Sequence of scene entries & 19k & 1.2K &  \tor{Scene intro, challenge, location}, and \tb{character descriptions} & Story \\
    
\textsc{TVStoryGen} \newline \cite{chen-gimpel-2022-leveraging}   & Crowd &  Fan wikis &  TV episode recap & 1.8k & 25.9K & \tor{Brief episode summary} and \tb{character wiki articles} & Scene Entry  \\
 \midrule 
 \multicolumn{2}{l}{\textbf{RPG Dialogue Datasets}} \\ \midrule 
 LIGHT \newline \cite{DBLP:conf/emnlp/UrbanekFKJHDRKS19} & Crowd & text game platform & Sequence of utterances &
 212 \newline (12 utt) & 276 & \tor{Location description}, \tb{persona statements} and \tb{held objects} & Dialogue \\
 TorchLight II~\cite{DBLP:conf/aiide/StegerenT20} & Professional & RPG Data & Sequence of quest stages with 0 or 1 utterances  & 157 \newline (3 utt) & 24 & \tor{Quest title, objectives, and details} 
 & Quest \\
 WoW~\cite{DBLP:conf/fdg/StegerenM21} & Professional & RPG data & NPC-uttered quest description & 61 \newline (1 utt) & 15 &  \tor{Quest title and objective}  & Quest \\
\name{} (Ours) &  Professional & RPG data & Complex quest dialogue tree
& 407 \newline (29 utt) & 1.3K & \tor{Quest title, objectives, and details}, \tor{walkthrough} and \tb{entity} wiki articles & Utterance %
      \\ \bottomrule
    \end{tabular}
    \caption{Comparison of \name{} to knowlege-constrained generation datasets. \name{} contains professionally-written complex dialogue trees from an actual
RPG, with more utterances, longer constraint passages, and higher-granularity annotations (of individual utterances) than other game dialogue datasets.}
    \label{tab:comparison}
    \endgroup
\end{table*}

\paragraph{Comparison with Related Datasets}
\label{sec:data-comparison}
\name{} is the first dataset to contain dialogue trees from an actual RPG annotated with game quest and biography specifications.
 \autoref{tab:comparison} compares our dataset to contemporaries with comparable input specifications and generation targets.  None contain all the components to target the complexity and granular specificity required to generate quest dialogues of the type found in \tow{}.
 \S\ref{app:comparison} provides an in-depth comparison with select works.

\subsection{Challenges}
Writing RPG-quality dialogue trees is difficult for human developers for its many interweaving considerations.
\name{} poses a similarily multi-faceted challenge to generation models:
\begin{itemize}
[itemsep=0ex,leftmargin=*]
    \item The tree must serve its \textbf{quest function}, containing input-specified player utterance options, NPC responses (including possibly specified emotions), and pieces of information the player must learn by the end. These NL specifications have highly diverse semantics (see \S\ref{app:quests}).
    \item The generated utterances must be \textbf{coherent},  \textbf{realistic}, and \textbf{engaging} to the player. 
    \item The NPC should embody the persona described in \textbf{their biography passage}, which describes personality, history, and relationships.
    \item To facilitate world building, the NPC should exposit details about \textbf{other entities} whenever it is contextually relevant, and should never \textbf{violate} the ontolology through contradiction. 
\end{itemize}
\vspace{-1mm}
Generating a 30-node (or larger) NPC dialogue tree while taking into account all of these criteria at once is a very difficult task, particularly given the shape of the branching, cycle-heavy tree structure. 

The average of 1321 constraining tokens and 406 utterance tokens %
poses a challenge to current NLP models, taking up e.g. half of the 4000-token context window of GPT-3 before factoring in other pieces of context such as few-shot examples.

\section{DialogueWriter Methods}
We introduce a set of neural methods for generating candidate utterances given the ontological specifications ($Q$, $B$, and $P$) from \S\ref{sec:task} and a partial subtree $S$ for a dialogue item in \name{}. We refer to these methods collectively as DialogueWriter models, which propose utterance nodes at a specified new location branching off the subtree.
Formally, given some ``most recent'' node $n \in S$, a DialogueWriter maps inputs $(Q, B, P, S, n)$ to a list of candidate utterances $[c_1, \dots, c_n]$ such that there is a directed edge $n \rightarrow c_i$.

\paragraph{Tree Linearization}
We consider language models (LMs) that accept linear input token sequences. We thus devise a traversal mechanism that converts a dialogue subtree into a maximal coverage linear history.
For ``most recent'' node $n$, we identify the longest possible path
from the start node to $n$, including cycles but only following a given edge once. This produces utterance history $H = [u_1, \dots, u_n]$. 
Exploring other tree encoding mechanisms, e.g. via graph encoders~\cite{banerjee-khapra-2019-graph, ouyang-etal-2021-dialogue} 
is left for future work.
With this linearization mechanism, we need only train a next utterance generator that conditions on linear history $H$. Methods to do so are explored below.

\subsection{Supervised Learning (SL) Models}
We fine-tune a T5-large sequence-to-sequence model~\cite{raffel-etal-2020-exploring} via supervised learning (SL) to generate $c_i$ given the concatenation $[B, Q, P, H]$.
We truncate context from the left of the when required given T5's 1024-token window, removing components of $B$ first.
We list last (and thus truncate last) the biographies of dialogue participants.
\S\ref{sec:app-prompts} shows example inputs.

\paragraph{Supervised Knowledge Selection (KS) Model}
We also train a version of the SL baseline that learns to decode support knowledge facts before conditionally generating the utterance $c_i$. 
This factorizes the next utterance generation into a two-step decision process: first selecting one or more facts from the provided knowledge constraints ($Q \cup B$), and second generating the utterance to reflect the selected facts. We thus make use of our node-level annotations to train the model to generate the concatenated sequence $[f_1^{(i)}, \dots, f_m^{(i)}, c_i]$ s.t. $ f_j^{(i)} \in Q \cup B$.

\subsection{In-Context Learning (ICL) Models}
As \knudge{} is relatively small, fine-tuning might not be effective at learning the difficult generation task. 
As such, we experiment with methods for \textit{in-context learning} (ICL) with OpenAI's \textit{text-davinci-003} GPT-3 model~\cite{DBLP:conf/nips/BrownMRSKDNSSAA20}. 
We inject $B$, $Q$, $P$, and $H$ into a formatted prompt that naturally elicits the next utterance as a continuation of $H$. \autoref{fig:method} depicts this process; full prompts are shown in \S\ref{sec:app-prompts}.
This creates a \textit{zero-shot} prompt.
When this does not fill out GPT-3's 4000-token window, we construct a \textit{few-shot} 
prompt by adding dialogs from training quests as exemplars, simulating a scenario in which a developer has written a partial set of quests and is working on a new one. We retrieve exemplars using Okapi-BM25~\cite{DBLP:journals/ipm/JonesWR00} with $[B, Q, P]$ as the query string.

\paragraph{ICL Knowledge Selection (KS) Model}
As with the SL framework, we also devise a version of the ICL DialogWriter that first decodes one or more support facts before generating an utterance. We elicit this behavior from GPT-3 by augmenting \textit{all} utterances in the dialogue history with support fact nodes, provided they exist (see \autoref{fig:cot-prompt}).
Further model training details are found in \S\ref{app:training}.
\section{Experiments}

\subsection{Baseline Models}
To measure the effect of conditioning on $Q$ and $B$, we compare against ablations to the ICL model: a \textbf{vanilla} model that conditions only on the participants $P$ and utterance history $H$, and a \textbf{quest only} model that conditions on $P$, $H$, and $Q$, but not $B$.

To measure the effect of node-level knowledge selection (\textbf{KS}), we compare against an ICL model that selects only \textbf{one} statement instead of many. We randomly sample gold facts to construct its prompt. 
We also compare against an \textbf{oracle} KS ICL model, which conditions on the full gold knowledge annotations for the reference utterance.
We maximize the number of in-context examples for all ICL ablations; e.g. the \textbf{vanilla} model's prompt can have dozens of such examples, as they are quite short. %
These ablations thus explore the tradeoff between the impact of the number of in-context examples and the presence of ontological statements.
\begin{table}[t!]
    \centering
    \footnotesize
    \setlength{\tabcolsep}{4pt}
    \begin{tabular}{llrrrrrr}
\toprule
& \multicolumn{2}{c}{\textbf{Gold}} & \multicolumn{2}{c}{\textbf{Bio} $B$} & \multicolumn{2}{c}{\textbf{Quest} $Q$} \\ \cmidrule(lr){2-3} \cmidrule(lr){4-5} \cmidrule(lr){6-7}
  $n = 1806$         &  \textbf{BL} &  \textbf{BS} &  \textbf{BL} &  \textbf{BS} &  \textbf{BL} &  \textbf{BS} \\
\midrule
\textit{Gold Reference} &        -- &           -- &       \textit{4.9} &           \textit{20.8} &       \textit{2.2} &           \textit{17.8} \\
        ICL-KS &        7.1 &            25.1 &       8.3 &           24.3 &       7.5 &           24.0 \\
    ICL-KS-One &        6.8 &            25.2 &       7.1 &           23.9 &       7.6 &           \textbf{24.5} \\
 ICL-KS-Oracle &        7.2 &            26.8 &       8.4 &           24.7 &       7.2 &           24.3 \\
           ICL &        \textbf{8.6} &            26.4 &       6.8 &           22.8 &       6.6 &           22.2 \\
ICL-Quest Only &        7.9 &            26.7 &       3.4 &           21.9 &       6.4 &           22.4 \\
       Vanilla &        6.8 &            \textbf{27.0} &       2.2 &           22.4 &       0.9 &           19.1 \\
         SL-KS &        2.6 &            21.3 &      \textbf{24.8} &           \textbf{26.6} &       9.3 &           21.3 \\
            SL &        2.9 &            23.5 &       7.4 &           24.0 &      \textbf{11.4} &           23.7 \\
\bottomrule
\end{tabular}
    \caption{NUP \textbf{BL}EU and \textbf{B}ert\textbf{S}core for models against gold utterances and statements in $B$ and $Q$. Results for the latter two shown beside the gold utterance's score.}
    \label{tab:auto}
\end{table}

\begin{table}[t!]
\footnotesize
\centering
\setlength{\tabcolsep}{2pt}
\begin{tabular}{lrrrr}
\toprule
{$n = 100$} &  Coherence &  Violation &   Using $B$ &   Using $Q$ \\
\midrule
\textit{Gold Reference}        &       \textit{3.94} &       \textit{3.97} &  \textit{3.50} &  \textit{3.45} \\

ICL-KS        &       \textbf{3.78} &       3.85 &  \textbf{3.29} &  \textbf{3.45} \\
ICL-KS-One    &       \textbf{3.73} &       3.80 &  \textbf{3.26} &  \textbf{3.45} \\
ICL-KS-Oracle &       \textbf{3.74} &       3.87 &  \textbf{3.23} &  \textbf{3.47} \\
ICL    &       \textbf{3.88} &       \textbf{3.97} &  \textbf{3.25} &  \textbf{3.43} \\
ICL-Quest Only &       \textbf{3.79} &       \textbf{3.90} &  3.03 &  3.21 \\
ICL–Vanilla         &       3.65 &       3.69 &  2.76 &  2.98 \\
SL-KS     &       2.52 &       3.85 &  2.17 &  2.09 \\
SL       &       2.70 &       3.74 &  2.35 &  2.38 \\
\bottomrule
\end{tabular}
\caption{NUP human evaluation results for in-context (ICL) and supervised learning (SL) DialogWriter methods with and without knowledge selection (KS).}
\label{tab:human}
\end{table}

\begin{figure*}[t!]
    \centering
    \scriptsize
    \begin{tabular}{p{0.97\textwidth}} \toprule
    \textbf{History:} \textit{> Agnes Needham: Please, you have to help me get my little Tucker back! He ran away and is going to get himself killed! Oh, I just know a raptidon is melting him with acid as we speak! > Player: Your child is missing? Where'd you last see him? > Agnes Needham: Oh, I just knew you were a good person. 'Agnes,' I said, 'this is the man to save your little Tuckie!' And I was right!} \\
    \textbf{Gold: Agnes Needham: He ran out into the wilderness a few days ago. I warned him about the raptidons, mantisaurs, and marauders - the toxic sulfur pools, and poisonous plants. But he didn't listen! } \\ \midrule
    \textbf{KS ICL:} Agnes Needham: He was always such \tor{a good boy}, but lately he's been acting so strange. I just don't understand why he would \tb{want to leave} \tor{Stellar Bay}. It's the only place he's ever known. \\
    \textbf{ICL:} Agnes Needham: He left a few weeks ago. Said \tb{he was going to Amber Heights} to join \tor{some revolutionary group}. I just know he's going to get himself killed if he's not careful! \\
    \textbf{ICL-Quest Only:} Agnes Needham: He left a few weeks ago, \tb{headed for Amber Heights}. He's going to get himself killed, I just know it!  \\
    \textbf{ICL-Vanilla:} Agnes Needham: He was playing in the street. But then he \tr{saw a sweet little kitty and chased it into the sewers!} Oh, the horror! \\
    
    \textbf{SL-KS:} Agnes Needham: \tr{If you're willing to pay handsomely,} I'll even \tb{pay you handsomely} if you convince him to return to me. \\
    \textbf{SL:} Player: I'll do everything I can. \\ 

    \bottomrule
    \end{tabular}
    \caption{[best viewed in color] Example generations by the in-context learning (ICL) DialogueWriter and supervised learning (SL) DialogueWriter. Models that condition on both biographical and quest passages are more likely to mention \tb{relevant quest information} while also \tor{referencing game entities and their backgrounds} without creating \tr{inconsistency, incoherence, or incongruity}. See  \S\ref{app:quests} and \S\ref{app:biographies} for full documentation of quests and entities referenced.}
    \label{fig:cherries}
    
\end{figure*}
\vspace{-1mm}
\subsection{Next Utterance Prediction (NUP)}
We split \knudge{} into train, development, and test splits on the basis of \textbf{quests} (28/5/12), such that at test time all input components will be unseen (test set $B$'s contain a combination of previously seen and totally novel entities). We evaluate utterances generated from a test set of 1800 partial trees.

\paragraph{Automatic Evaluation}
We use reference-based metrics BLEU-4 \cite{papineni2002bleu} and
 re-scaled BERTScore-F1 \cite{DBLP:conf/iclr/ZhangKWWA20}.
We evaluate generated next utterance against the following single- and multi-reference sets: 1) the gold utterance $n_i$,  2) the quest statements in $Q$, and 3) the biography statements in $B$. 
We evaluate the \textbf{gold} utterance to show a performance upper bound. 

\paragraph{Human Evaluation}
\label{sec:human-eval}
In coordination with a data specialist, we conducted human evaluation to examine models' qualitative NUP performance.
Generated utterances over 100 test items were judged on a 4-point Likert scale for each of four criteria:
\vspace{-2mm}
\begin{enumerate}[wide, labelindent=0pt,itemsep=0ex]
    \item \textbf{Coherence:} does the utterance follow naturally from the utterances in the history? 
    \item \textbf{(Non-)Violation:} does the utterance create contradictions with any of the sentences in the biographical or quest passages? 
    \item \textbf{Biography Usage}: does the utterance \textit{make use} of the input biographical passages in $B$? 
    \item \textbf{Quest Usage}: does the utterance progress the dialogue according to the quest sentences in $Q$? 
\end{enumerate}

We provide the full set of annotator instructions with guidance for each Likert scores in \S\ref{app:human-eval}. Results were verified via bootstrap testing.

\subsection{Full Dialogue Generation Case Study}
\label{sec:tree-eval}
We run a case study for a full tree generation scenario. We task models with generating 10 rounds of dialogue given just the specifications $B$, $P$, $Q$, and one starting utterance.\footnote{This serves to specify utterance format in cases of zero-shot learning, which is the case for all designer-written items.} 
At each turn, we generate three candidate nodes using the writer, then randomly ``commit'' one to the linear history. This creates a 31-node tree (example in 
\autoref{fig:tree_eval_afm}
) that can serve as a `spine' of proposed content to be fleshed out by a developer into a more complicated tree. 
We hand-selected a set of 8 test dialogues from the game with varying quest roles, e.g. starting, continuing, and ending quests. Additionally, we had a professional game designer write specifications for \textbf{2 totally novel quests} with overlapping entities from the original game; from these we constructed 8 more test items.

We generated trees using the KS, regular, quest only, and vanilla ICL DialogueWriters, then showed them to data specialists for evaluation following the ACUTE-Eval~\cite{li-etal-2019-acute} protocol of pairwise comparison.
Annotators were asked to select which of two trees were preferred for the following criteria: \textbf{coherence}, \textbf{nonviolation}, \textbf{biography} and \textbf{quest} usage analogous to \S\ref{sec:human-eval}, and additionally
\vspace{-1mm}
\begin{enumerate}[wide, labelindent=0pt,itemsep=0ex]
\setcounter{enumi}{4}
\item \textbf{Content Suggestion}: Do the multiple candidates at each turn propose interesting subtrees?
\item \textbf{Engagingness}: does the tree hold your attention and make you want to hear more from the NPC?
\end{enumerate}

\section{Results and Discussion}
\textbf{Next utterance prediction} results under automatic and human metrics are shown in \autoref{tab:auto} and \autoref{tab:human}.
We note that automatic metrics for generation that check for lexical or semantic overlap with a set of references is not directly suited for evaluating generations in \knudge{}; this can be seen from the extremely low performance of \textit{the gold utterances themselves} under these metrics. We find that professionally written utterances do not always have high overlap with knowledge statements themselves. Gold utterances also do not score perfectly under utterance-level human evaluation of $Q$ and $B$ usage, as not every real-world utterance reflects the ontology. It might instead provide other qualities like realism and fluency. This highlights the challenge of identifying and then pursuing desiderata of RPG-quality dialogue trees.

We observe that the best-performing model under overlap metrics with $B$ and $Q$ are the T5-based SL models. This reflects that these models have learned to copy spans directly from the context into the generation, hence scoring highly on ngram-based overlap with it while scoring poorly on gold utterance overlap and human evaluation.

\autoref{tab:auto} depicts the trend that KS variants of the ICL model score a point or two higher than non-KS on overlap with $B$ and $Q$, reflecting that KS effectively selects and cues the infusion of specific facts into generations.
Oracle KS improves BLEU score with $B$ but not $Q$, while one-fact KS has the opposite effect. The ICL ablations of $B$ and of $\{B, Q\}$ have according drops in overlap with both fact sets. The \textbf{Quest}-only ICL DialogueWriter can generate coherent utterances that reflect the quest specifications; however, as illustrated by \autoref{tab:human} and \autoref{fig:cherries}, this can sacrifice lore references.

\autoref{tab:human} also shows that all fully $B$ and $Q$-conditioned ICL models perform equivalently under all human metrics except (non)violation, which KS models perform a decimal point worse at. We can conclude that KS improves the capacity of ICL writers to directly reflect knowledge passages (i.e. by copying spans), at the expense of a slightly higher likelihood of contradictions. This characterization can be appealing to a game developer; they might prefer for the automatic writer to use their own provided wordings of various facts when generating candidates. We note that no model approaches gold performance at reflecting $B$.
\vspace{-1mm}
\paragraph{Qualitative Analysis}
\autoref{fig:cherries} depicts example outputs by models on an NUP example. We highlight cases in which the models succeed at the desiderata that we strive for in \name{}: to convey quest and lore specifications naturally through the interaction. However, we see that SL models and ablated ICL models are less successful.
We observe that the gold utterance is more infused with desirable information than any generation; it references the quest's next location and numerous adversaries that the player will run into, while effectively reflecting the NPC's overprotective parent persona.
This highlights a performance gap between neural and human writers to be addressed by future work. 
We note that not reaching human performance does not preclude DialogueWriters from being useful to writers, as they can still be used to suggest new directions for dialogues to be verified or modified in a human/AI collaborative writing process.
\begin{table}[t!]
    \centering
    \footnotesize
    \setlength{\tabcolsep}{2.3pt}
    \begin{tabular}{lrrrrrr}
\toprule
$n = 16$ &  Coh. &  Viol. &  Use $B$ &  Use $Q$ &  Content &  Engaging \\
\midrule
ICL-KS        &       50.0 &       68.8 &       68.8 &        \textbf{75.0} &     43.8 &          37.5 \\
ICL    &       \textbf{81.2} &       \textbf{75.0} &       \textbf{81.2} &        \textbf{75.0} &     \textbf{75.0} &          \textbf{75.0} \\
ICL-Quest &       56.2 &       56.2 &       43.8 &        50.0 &     50.0 &          68.8 \\
ICL-Vanilla         &       12.5 &        0.0 &        6.2 &         0.0 &     31.2 &          18.8 \\
\bottomrule
\end{tabular}
    \caption{Pairwise comparison (\% head-to-head wins) between generated trees from ICL DialogueWriters. For example, 81.2\% of the time ICL outputs were preferred over a competing approach with respect to coherence.}
    \label{tab:tree_eval}
\end{table}

\textbf{Full dialogue generation case study} results are found in \autoref{tab:tree_eval}. Model performance is measured as the rate at which annotators selected its generated tree in a pairwise comparison under the 6 criteria listed in \S\ref{sec:tree-eval}.
We find that annotators preferred the trees of the ICL writer most frequently compared to the other models under all criteria except $Q$ usage. 
We also find that the ICL-KS method is more frequently preferred for the never-before-seen quests written by the professional designer, while it is less frequently preferred for dialogues from the actual game. The opposite holds for ICL-Quest only.
A possible explanation is that the designer provided many biographical specifications for his quests (enough to fill out the GPT3 context window), without which there might not be a sufficient signal for the Quest only model to generate desirable utterances. 
Another reason might be that the few-shot learning benefits of the Quest only model not conditioning on $B$ are lessened by the domain shift from \name{} to the designer's writing.

\section{Related Work}

\citet{DBLP:conf/sigdial/SiAR21} focus on the task of story continuation through dialogue between multiple characters while modeling the inter-character relations. However, such past work does not concern with the notion of grounding knowledge or quest objectives to be covered in the generated dialog.  
\citet{DBLP:conf/aiide/StegerenT20} propose three sources for building NPC dialogue corpora. However, their proposed datasets do not contain any grounding annotation and are not accompanied by explicit descriptions of entities and characters. %
\citet{DBLP:journals/corr/abs-2210-07109} explore automatic generation of conversational turns by players of the tabletop RPG, Dungeons and Dragons, 
in which NPCs serve a very different role in the gameplay.
Scheherazade’s Tavern \cite{DBLP:conf/fdg/AljammazOWM20} augments a pattern-matching-based NPC interaction system with facts the character knows about the game world.
More broadly, past work has explored applications of text generation in various gaming applications such as quest description generation \cite{DBLP:conf/fdg/StegerenM21}, dialogue generation \cite{DBLP:conf/sigdial/SiAR21},  persona-specific agents in text environments \cite{DBLP:conf/emnlp/UrbanekFKJHDRKS19}, and new text world generation \cite{DBLP:conf/aaai/FanURDQKPKRSW20,DBLP:conf/acl/AmmanabroluJR22}.

Past work has pursued dialogue systems that steer the conversation towards a topic \cite{DBLP:conf/acl/WuGZWZLW19} or a given NL sentence \cite{DBLP:conf/acl/SevegnaniHKR20, DBLP:conf/naacl/GuptaJB22} while conversing with a user. Other work in NLG has explored generating outputs with high-level NL specifications such as string item agendas \cite{DBLP:conf/emnlp/KiddonZC16}, sets of facts \cite{DBLP:conf/coling/OrbachG20}, or author goals \cite{DBLP:conf/aaaiss/Riedl09}. \name{} also comprises NL specifications, though they are comparably richer.

\section{Conclusion}
Humans play games to be entertained, and they pay money expecting a high-quality experience.  When a game requires dialogue to advance a carefully crafted storyline, this should be both engaging as well as consistent with the larger narrative. Language models are increasingly capable of producing engaging dialogues, and researchers have been exploring how to ensure that such content is consistent with underlying knowledge specifications.  To date, this research has focused on scenarios developed for the sake of experimentation, rather than actual high-quality game data.

In this paper, we introduce \name{}, a dataset of NPC dialogue trees coupled with a relevant game ontology, drawn from the title \emph{The Outer Worlds}.
In contrast to prior work, \name{} is based on content created by a triple-A game (i.e. high-budget, high-profile game) development studio, %
thereby exemplifying real-world complexities in NPC dialogue authoring. 
We illustrate that LMs are indeed able to generate fluent dialogues that relate to provided game lore.  However, the straightforward application of such technologies does not match the quality of professional game writers.
We hope that \name{} aids the development and evaluation of new techniques for faithful game dialogue generation.

\section*{Limitations}
One limitation of our work is that our considered models use simplifications to model the complex branching trees by linearizing the nodes. We moreover only evaluate next-utterance prediction and relatively simple end-to-end dialogue trees which do not approach the complexity of the actual trees. 
More can be done on this front in future work to develop methods that generate more complicated tree structures. We also leave for future work the evaluation of whether end-to-end generated dialogues ``check all the boxes'' of quest requirements, completing \textit{all} user-provided specifications.

We also find that the proposed DialogueWriter models leave room for improvement on persona embodiment. Human-quality utterances more seamlessly and dynamically incorporate emotions fitting of characters and situations, while model-generated utterances can be comparatively bland. This work also focuses on \textit{side quests} whose NPCs are generally not as fleshed out as those in main quests. Generating quests containing major NPCs with long bios and important roles in the main story of a game, e.g. companion characters, is also left for future work.

\name{} recasts a set of fan articles about an existing game as specifications to an automatic dialogue tree writer. It therefore
assumes that game developer will write structured game lore and high-level quest specifications in a similar manner \textbf{beforehand} when coming up with new content. Thus, we provide only a partial solution to the problem of automatic NPC dialogue generation, and future work can look at copilot tools for authoring such high-level quest specifications and design of new characters.

We report results with large pre-trained language models. It is difficult to know whether the game data used for experimentation is part of the training data for such models, as \tow{} came out in 2019. As such, the results from such large language models should be interpreted with caution. We partially mitigate the issue by having an expert game developer construct a totally new quest specification, and report results on the this previously 'unseen' test data. 

\section*{Ethics Statement}
We acknowledge that there maybe bias in the data used to train the neural language models considered in this paper (T5 and GPT-3) that would lead to NPC dialogues that are offensive, implicitly or explicity discriminatory. 
This poses a potential risk for deployed models, as using the proposed DialogueWriters as content suggestion tools might lead to RPG content that reflects these biases. We hope that professional game developers will have the resources to moderate damaging content before it makes its way into released products.

\bibliography{anthology,custom}
\bibliographystyle{acl_natbib}

\appendix

\begin{figure*}[t]
    \centering
    \includegraphics[width=.90\textwidth]{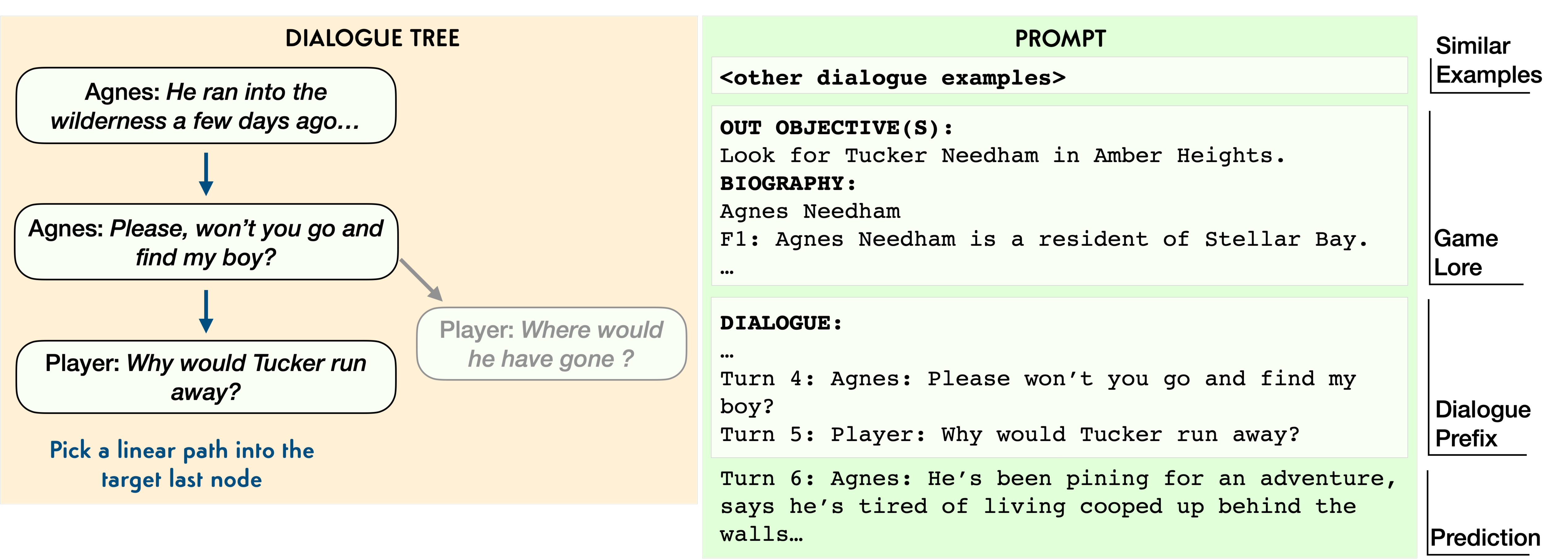}
    \caption{Overview of our method for constructing in-context learning prompts from constraints and dialogue  history.}
    \label{fig:method}
\end{figure*}
\begin{figure*}[t!]
    \centering
    \includegraphics[width=\textwidth]{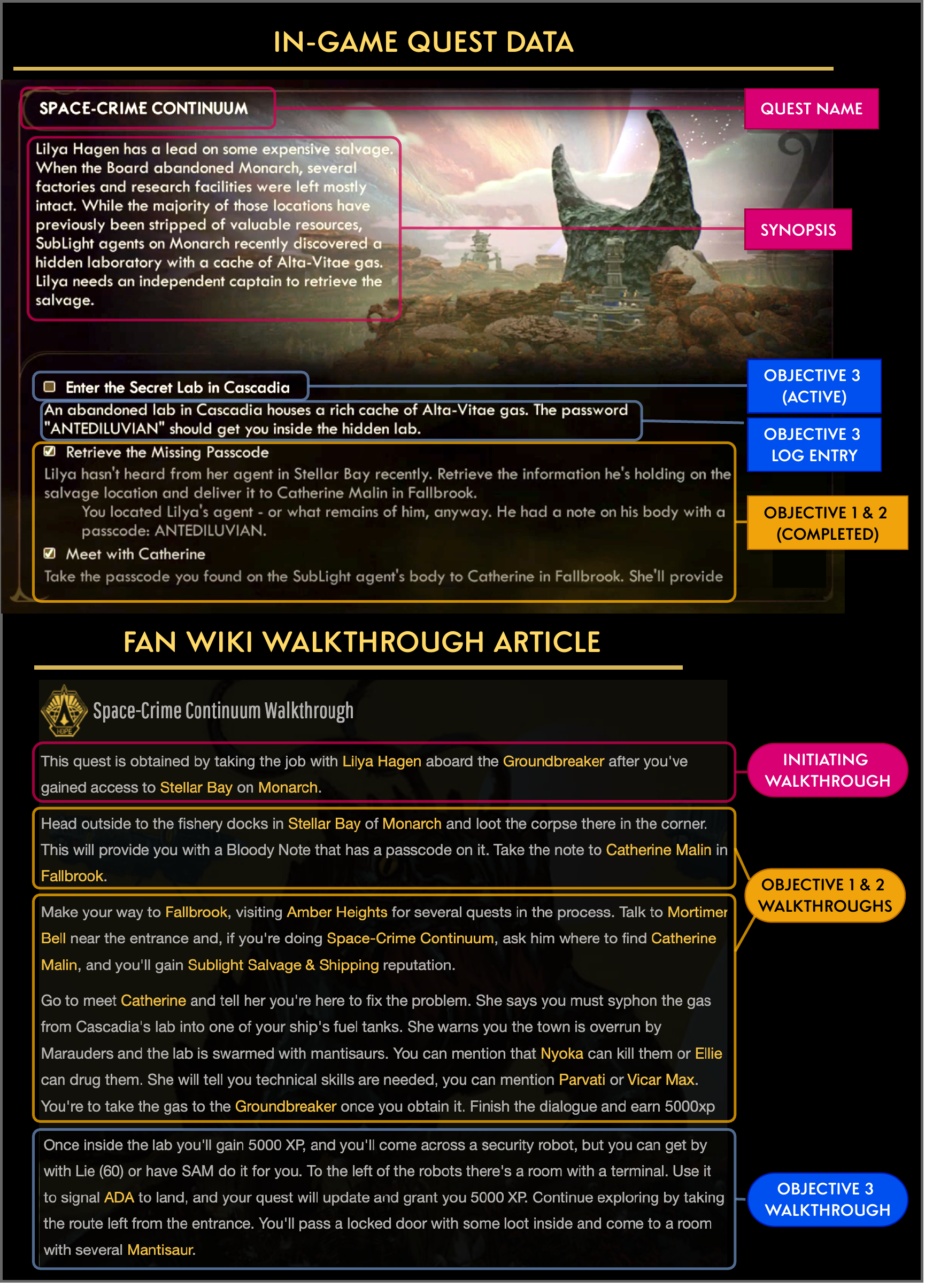}
    \caption{Anatomy of a \name{} quest. At any time, an \ow{} quest has currently \textcolor{blue}{active} and previously \textcolor{orange}{completed} objectives.  To construct its \name{} representation, the quest's high-level \textcolor{magenta}{synopsis}, objectives, and associated log entries from the game data are annotated with corresponding walkthrough article passages.}
    \label{fig:quest_anatomy}
\end{figure*}
\section*{Appendix}

\section{Cost of Authoring NPC Dialogues}
\label{app:cost}
Outer Worlds has 10 narrative/design credits\footnote{\url{https://www.imdb.com/title/tt9417446/fullcredits}}, which seems to be about average (Fallout has 4-12, while Skyrim has 9). Per salary.com, that's a position with an average salary of \$58k per year (and probably more for AAA titles). Given a 2-3 year average development time for AAA titles, that works out to a conservative ballpark estimate of \$1.2m for just this one game.  

\section{Dataset Construction Details}
\label{app:dataset}
\subsection{Data Sources}
Quest data and walkthrough passages were pulled from the \ow{} wiki of Fextralife,\footnote{\url{https://www.theouterworlds.wiki.fextralife.com}} a gamer-focused site containing fan-made walkthroughs for many popular RPGs. 
Game entity biographies were collected from Fandom.\footnote{\url{https://theouterworlds.fandom.com}

}
The biography passage for a given entity is the same across all quests in which the entity appears in \name{}, and the set of entities is the same for all dialogues in a quest.
Passages were segmented into individual sentences via punctuation boundaries.
We identified relevant dialogues and their decision points using playthrough videos by the YouTube user, 
LordMatrim.\footnote{\url{https://www.youtube.com/@l0rdmatrim}} All wiki articles were written in English by site users. 
\section{Quest Anatomy and Example Items}
\label{app:quests}

\autoref{fig:quest_anatomy} provides a detailed anatomy of a \name{} quest, combining in-game quest data with corresponding passages from the fan walkthrough.
\autoref{fig:ex-quests} 
shows example quest items with corresponding game data and walkthrough passages segmented into statements.
\begin{figure*}[]
    \centering
    \scriptsize
    \begin{tabular}{p{.95\textwidth}}
    \toprule
\textbf{Quest Name:} \textbf{A Family Matter}\\
\textbf{Synopsis:} [0] Tucker Needham ran away from Stellar Bay a few weeks ago to join the Iconoclasts in Amber Heights. [1] His mother Agnes is willing to pay handsomely if you can locate her son and convince him to return\\
\textbf{Walkthrough:} \textit{[0] You can begin this quest by talking to Agnes Needham in Stellar Bay, Monarch. [1] Agnes is by the town's south-east exit, visibly shaken and calling for help. Hear her out and offer to find her son to being the quest.}\\
\midrule
\textbf{Objective 1:} \textbf{Look for Tucker Needham in Amber Heights}\\
\textbf{Game Log:} [0] Amber Heights is the settlement that houses the Iconoclasts on Monarch. [1] If Tucker Needham survived his travels, his mother thinks he'll be there.\\
\textbf{Walkthrough:} \textit{[0] Head South from Stellar Bay and follow the east road. It will take you to Amber Heights. [1] Head up the hill and go into a residence on the left to meet Tucker Needham.}\\
\midrule
\textbf{Objective 2:} \textbf{Convince Tucker to Return Home}\\
\textbf{Game Log:} [0] Now that you've found Tucker Needham in Amber Heights, convince him to return home to his mother in Stellar Bay.\\
\textbf{Walkthrough:} \textit{[0] Introduce yourself, and then you can mention your surprise that this grown man is the "little boy" that ran away. You'll earn 7500xp [1] Explain to him that she made it sound as if he was a boy in danger, [2] and he'll say she has been overprotective all her life, [3] and he is ready to live his life without her protection. [4] You can persuade (55) or intimidate (55) to expedite things and get him to go back, [5] or you can ask him what he wants to do about it. [6] The last option will have him tell you to report that he is dead. [7] You can express your concern about what that will do to Agnes, [8] and then either ask for something that would prove a body, or reject the proposition. [9] If you persuade him to go back, you'll get 7500xp and can return to Stellar Bay to see things play out.}\\
\midrule
\textbf{Objective 3:} \textbf{Return to Agnes Needham in Stellar Bay}\\
\textbf{Game Log:} [0] You convinced Tucker Needham to return home to Stellar Bay. [1] Agnes promised a reward for bringing her son back.\\
\textbf{Walkthrough:} \textit{[0] You'll find his mother is still condescending to him, [1] and you can help him by saying he's a grown man. [2] You'll get 7500xp. [3] If you stick around and talk to them some more you'll see Tucker is standing up for himself. [4] You'll receive 625 Bit Cartridge, Monarch Stellar Industries Reputation and 15000xp.}\\
\midrule
\midrule
\textbf{Quest Name:} \textbf{The Commuter}\\
\textbf{Synopsis:} [0] The Iconoclasts are due to receive a shipment of vital supplies from Carlotta, a sympathizer that resides in Stellar Bay. [1] The meeting is set to occur at the Bayside Terrace warehouse.\\
\textbf{Walkthrough:} \textit{[0] The quest can be obtained by asking Graham if there is anything that needs doing. [1] He is trying to get an old printing press running, but the replacement rollers he'd requisitioned haven't arrived yet. [2] They were supposed to be delivered by Huxley, but she is still recovering and unable to make the delivery. [3] Graham asks the player to meet the supplier in her stead, and to pick up high-capacity data cartridges with the funds left over from the previous shipment. [4] Zora will interject to ask the player to buy food and medicine instead with the leftover money.}\\
\midrule
\textbf{Objective 1:} \textbf{Get the Printing Press Rollers from Carlotta}\\
\textbf{Game Log:} [0] Travel to the warehouse at Bayside Terrace and find Graham's contact, Carlotta. [1] She should have a shipment for him. Retrieve it. [2] Speak to Carlotta\\
\textbf{Walkthrough:} \textit{[0] Clear out the Sublight squad that is hunting Carlotta [1] Carlotta is behind a locked door to the east. [2] Activate the intercom next to the door to speak to her and she will unlock it. [3] Go inside and speak to her again to obtain the rollers needed to complete the quest, then choose between the high-capacity data cartridges or food and medicine.}\\
\midrule
\textbf{Objective 2:} \textbf{Get High-Capacity Cartridges or Extra Supplies from Carlotta}\\
\textbf{Game Log:} [0] Graham wants to tack on some high-capacity cartridges to his order, but Zora would prefer it they could get extra food and medical supplies. [1] You got extra supplies for Zora (or) You got High-Capacity Data Cartridges for Graham.\\
\textbf{Walkthrough:} \textit{}\\
\midrule
\textbf{Objective 3:} \textbf{Return to Graham}\\
\textbf{Game Log:} [0] Bring the needed parts back to Graham at Amber Heights\\
\textbf{Walkthrough:} \textit{[0] Return to Graham and you'll find him arguing with Zora about the Van Noys, a unit of the Iconoclasts that is MIA. [1] Inform Graham that you got his rollers, and food and medicine if that was your choice. [2] You'll receive 7500xp and Zora will ask when the next drop is. [3] Inform her that Sanjar has made it illegal to trade with the Iconoclasts.}\\
\midrule
\midrule
\textbf{Quest Name:} \textbf{Who Goes There}\\
\textbf{Synopsis:} [0] The Groundbreaker's Mardets have a bounty for a criminal on the run in the Groundbreaker's Back Bays. [1] You've agreed to hunt down the unlawful Captain Gunnar MacRedd. [2] Return his lighter to Commandant Sanita to claim the bounty.\\
\textbf{Walkthrough:} \textit{[0] This quest is obtained at Groundbreaker, [1] by speaking to Comdt. Sanita or perusing the bounty board}\\
\midrule
\textbf{Objective 1:} \textbf{Hunt Down and Kill Captain McRedd}\\
\textbf{Game Log:} [0] Based on the bounty listing, Captain McRedd was last sighted in the Back Bays. [1] Head there and take him out.\\
\textbf{Walkthrough:} \textit{[0] You can find Captain MacRedd in the Back Bays area of the Groundbreaker. [1] To get there head down the elevator in the promenade, [2] and you can't miss him. [3] You can pass a Persuade (40) check to get him to put his gun down, [4] otherwise you'll have to kill him and all his guards. [5] If you kill him he drops the Unique Weapon: Montag. [6] You'll get 6000xp and MacRedd's Lighter. [7] If you persuaded him, use Perception to note it says "Sanita" on the lighter. [8] MacRedd will mention it was given to him by Sanita in remembrance of a 'carnal understanding' they had a few years back.}\\
\midrule
\textbf{Objective 2:} \textbf{Claim the Bounty's Reward from Comdt. Sanita}\\
\textbf{Game Log:} [0] McRedd gave you his lucky lighter to give to Sanita. [1] Go turn it in to resolve his bounty.\\
\textbf{Walkthrough:} \textit{[0] Turn the lighter in to Commandant Sanita to claim the bounty.}\\
\midrule
\bottomrule
    \end{tabular}
    \caption{Example Quest Items}
    \label{fig:ex-quests}
\end{figure*}

\section{Example Entity Biography Passages}
\label{app:biographies}
\autoref{tab:entities}
shows example entities from \tow{} with corresponding biographical passages.
\begin{figure*}[]
    \centering
    \scriptsize
\begin{tabular}{p{.95\textwidth}}
\toprule
\textbf{Entity:} \textbf{Agnes Needham}\\
\textbf{Appears in:} \textit{A Family Matter}\\
\textbf{Bio:} [0] Agnes Needham is a resident of Stellar Bay and the mother of Tucker Needham. [1] Agnes' overprotective style of mothering has led her son, Tucker Needham, to run away from home so he can experience life. [2] Despite Tucker being 42 years old, she still thinks of him as her 'little boy'. [3] You can find her by Stellar Bay's south-east exit, visibly shaken and calling for help.\\
\midrule
\textbf{Entity:} \textbf{Tucker Needham}\\
\textbf{Appears in:} \textit{A Family Matter}\\
\textbf{Bio:} [0] Tucker Needham is a former resident of Stellar Bay who left to join the Iconoclasts. [1] Before the quest A Family Matter, he can be found in Amber Heights. [2] Tucker was coddled by his mother from a very young age, [3] the latter insisting that danger lurked around every corner on Monarch. [4] His mother's overprotectiveness extended well into Tucker's adulthood, [5] leading him to seek to be free in any way possible. [6] After hearing Graham Bryant's broadcasts, Tucker left Stellar Bay to be truly free by joining the Iconoclasts at Amber Heights. [7] He is dazzled by Graham's preachings on true unfettered freedom from the corporate way of life and attributes his enthusiasm to his 'childhood trauma'. [8] He is willing to do anything to remain free, even faking his own death to prevent his mother from continuing to send people to look for him.\\
\midrule
\textbf{Entity:} \textbf{Raptidon}\\
\textbf{Appears in:} \textit{A Family Matter}, \textit{At Central}, \textit{Bolt With His Name}, \textit{Journey Into Smoke}, \textit{Makes Space Suits Wont Travel}, \textit{The Amateur Alchemist}, \textit{The Distress Signal}, \textit{The Doom That Came To Roseway}, \textit{Vulcans Hammer}\\
\textbf{Bio:} [0] Raptidons are giant cat/reptile-like creatures that inhabit various planets in Halcyon. [1] They are creatures native to Monarch. [2] however some corporations have illegally imported them to other planets, [3] such as Auntie Cleo who relocated a group of them to Roseway. [4] Raptidons are of corporate interest due to their potential for producing new chemical by-products which, [5] when refined, can be used to create new board approved products.\\
\midrule
\textbf{Entity:} \textbf{Sulfur Pits}\\
\textbf{Appears in:} \textit{A Family Matter}\\
\textbf{Bio:} [0] The Sulfur Pits are a point of interest on the western side of Monarch. [1] They are located southwest of Terra One Publications and directly northeast of the Gunship Crash Site. [2] The Sulfur Pits have a large variety of Raptidons and many deceased marauders. [3] The area consists largely of Sulfur Pits. [4] When an entity comes in contact with a sulfur pit, [5] they receive the acid effect for the duration of touching the pit.\\
\midrule
\textbf{Entity:} \textbf{Monarch}\\
\textbf{Appears in:} \textit{A Cysty-Dance With Death}, \textit{A Family Matter}, \textit{Bolt With His Name}, \textit{Little Memento}, \textit{Makes Space Suits Wont Travel}, \textit{Mandibles Of Doom}, \textit{Slaughterhouse Clive}, \textit{Space-Crime Continuum}\\
\textbf{Bio:} [0] Monarch, previously known as Terra 1, is one of the many moons of the gas giant Olympus and the site of a failed colony. [1] Terra 1 was initially designated as the primary colonization target of the Halcyon system. [2] The Halcyon Holdings Corporate Board had intended to completely terraform the moon, [3] wiping out the local fauna and flora and replacing it with plants and wildlife native to Earth. [4] However, the terraforming process unexpectedly caused the native species to mutate and grow to significantly larger sizes, [5] rendering them more dangerous and severely crippling the colonization effort. [6] Due to the hostile environment which they had created, [7] the Board was forced to enact a Hazard Clause covering the entirety of Terra 1. [8] Public notice of the clause's issuance was sent to everyone operating on Terra 1 and led to the evacuation of almost all corporations from the moon. [9] However, one corporation took advantage of the chaos of the evacuation to exploit a legal loophole which allowed them to, [10] as the last corporation remaining on the planet, [11] acquire the planet from the Board. [12] This corporation, under the leadership of Sanjar Nandi and Graham Bryant subsequently rebranded itself to Monarch Stellar Industries (MSI), [13] in line with the renaming of the planet to 'Monarch'. [14] The actions of MSI earned them the ire of the Board, [15] who retaliated by effectively placing the moon under indefinite embargo, [16] refusing to allow legal transit either in or out. [17] the Board aggressively spread propaganda about Monarch to convince the rest of the population that it was both uninhabited and uninhabitable. [18] This has greatly hampered MSI's attempts to be recognized as a legitimate corporation and is a thorn in the side of its CEO, Sanjar Nandi. [19] Monarch also has an ocean which goes around the moon at the "twilight band". [20] It is where the colonists and Monarch Stellar Industries farm their saltuna.\\
\midrule
\textbf{Entity:} \textbf{Stellar Bay}\\
\textbf{Appears in:} \textit{A Family Matter}, \textit{Bolt With His Name}, \textit{Canids Cradle}, \textit{Flowers For Sebastian}, \textit{Herricks Handiwork}, \textit{Mr Picketts Biggest Game}, \textit{Passion Pills}, \textit{The Stainless Steel Rat}\\
\textbf{Bio:} [0] Outside the city walls, the lands were overrun by the native wildlife, as well as marauders and outlaws. [1] Stellar Bay is a company town located on the planet Monarch. It is owned and operated by Monarch Stellar Industries. [2] Stellar Bay is the largest saltuna producer on the Halcyon colony and used to be one of the most important suppliers of this resource.\\
\midrule
\textbf{Entity:} \textbf{Fallbrook}\\
\textbf{Appears in:} \textit{A Cysty-Dance With Death}, \textit{Slaughterhouse Clive}, \textit{Space-Crime Continuum}, \textit{Spratkings}\\
\textbf{Bio:} [0] Fallbrook is a company town located on Monarch, [1] loosely run by the SubLight Salvage and Shipping Corporation. [2] Fallbrook is a small town built into the side of a mountain, [3] whose construction was masterminded by Catherine Malin. [4] Fallbrook has a lot of activities to offer to its visitors, [5] from those who search for activities of leisure to those with proclivities for vice.\\
\midrule
\textbf{Entity:} \textbf{Cascadia}\\
\textbf{Appears in:} \textit{Space-Crime Continuum}, \textit{The Chimerists Last Experiment}, \textit{The Ice Palace}\\
\textbf{Bio:} [0] Cascadia is an abandoned company town that was owned and operated by Rizzo's before it withdrew from Monarch. [1] It is now used as a stronghold by the Marauders. [2] The main attraction is the Cascadia Bottling Plant and, [3] for those in the know, [4] the Rizzo Secret Laboratory hidden underneath the Rizzo Sweets Shoppe.\\
\midrule
\textbf{Entity:} \textbf{Amber Heights}\\
\textbf{Appears in:} \textit{Little Memento}, \textit{Odd Jobs}, \textit{Sucker Bait}, \textit{The Commuter}\\
\textbf{Bio:} [0] Amber Heights is a location in the Monarch Wilderness and the base of operations for the Iconoclasts. [1] The Iconoclasts run the place somewhat like a commune. [2] Amber Heights was once the place of residence of the entire executive dome of Monarch Stellar Industries. [3] It is now in ruins after a massacre in the past. [4] They lived there with their families and it was the company's operations center on Monarch. [5] Just after The Board approved the evacuation of the planet through the Hazard Clause, Amber Heights was besieged by a gang of pirates who ransacked the town and massacred all its inhabitants. [6] This tragedy was known as "The Amber Heights Massacre". [7] They were secretly assisted by MSI employee, Graham Bryant, who believed that the massacre would aid him in his quest to rid the colony of corporate influence. [8] In 2345, the same Graham Bryant formed the Iconoclasts and settled the group in the deserted town.\\
\bottomrule
\end{tabular}

    \caption{Example entity biographies that appear as constraining knowledge in \name{} quest dialogs}
    \label{tab:entities}
\end{figure*}

\section{Example Dialogues Items}
\label{app:dialogues}
\begin{figure*}
    \centering
    \includegraphics[width=\textwidth]{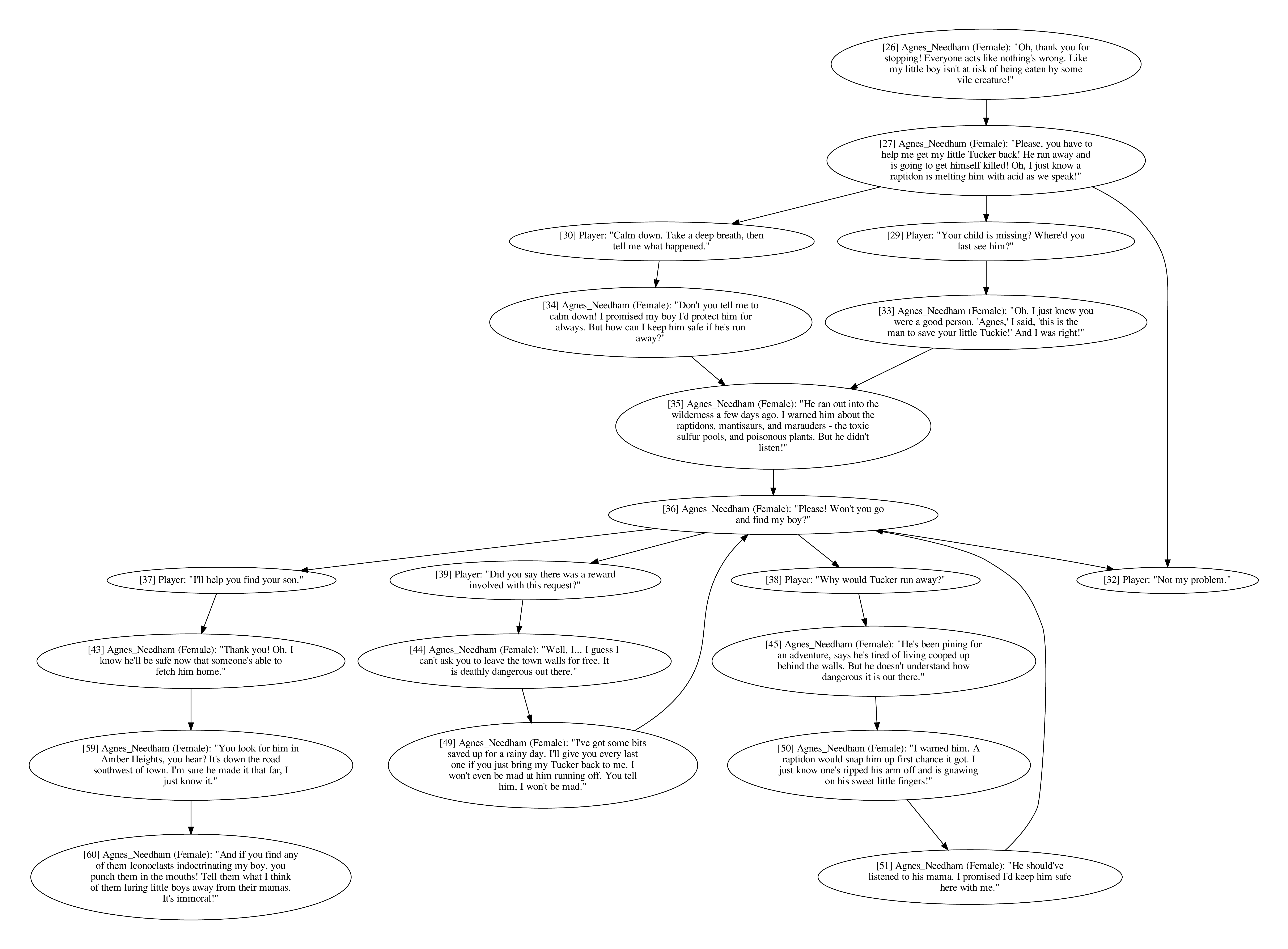}
    \caption{Full dialogue tree in \name{} for motivating example in \autoref{fig:overview}.}
    \label{fig:family-matter-tree}
\end{figure*}

\begin{figure*}
     \centering
    \includegraphics[width=\textwidth]{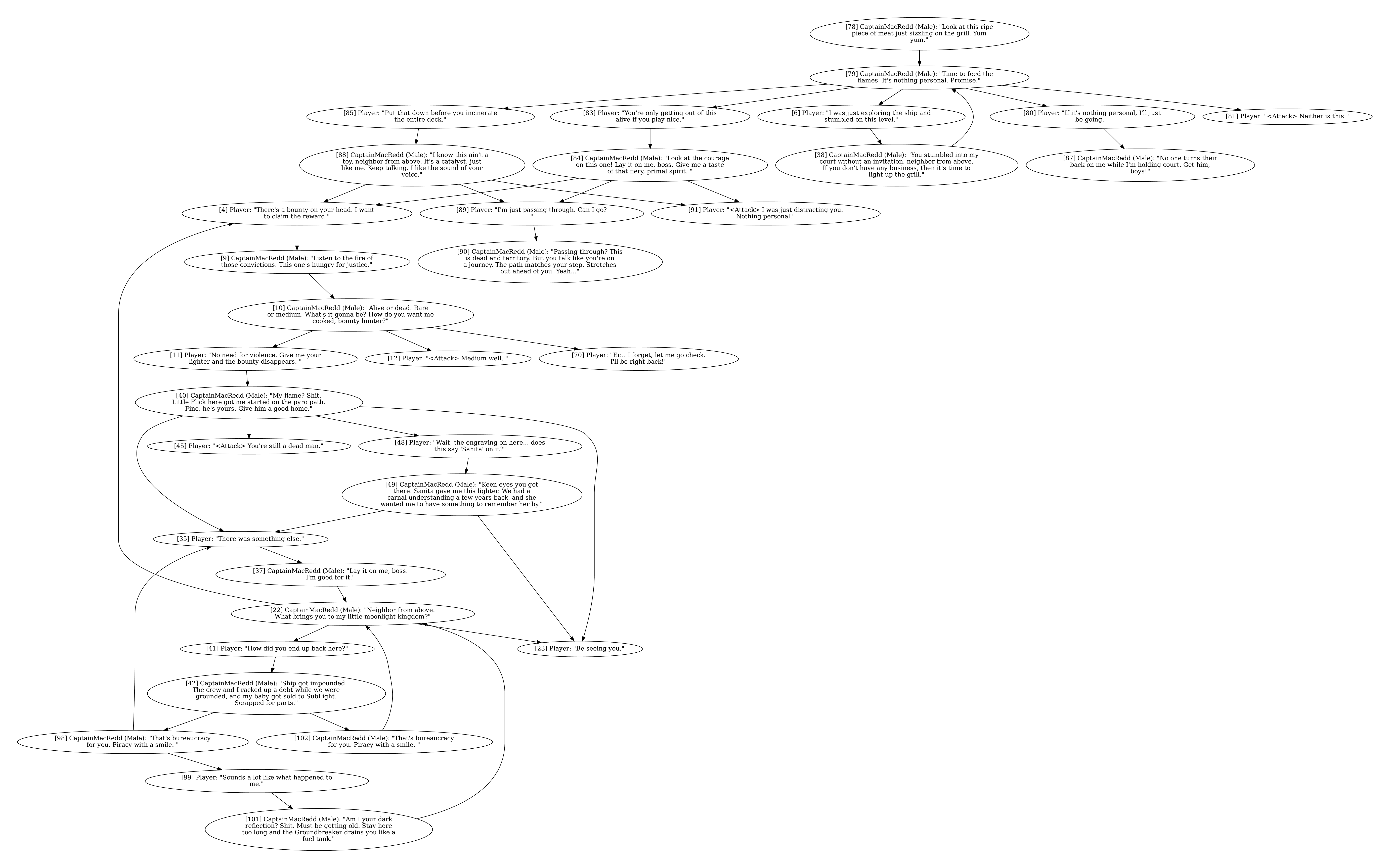}
    \caption{Example full dialogue tree for dialogue \textit{who\_goes\_there\_01} in \name{}.}
     \label{fig:tree-2}
\end{figure*}

\autoref{fig:family-matter-input} depicts a full example input item conveying quest, biographical, and participant specifications.
 \autoref{fig:family-matter-tree}, \autoref{fig:tree-2}, and \autoref{fig:tree-3} depict example dialogue trees.

\begin{figure*}[]
    \centering
    \scriptsize
\begin{tabular}{p{.95\textwidth}} \toprule
\textbf{Dialog:} \textit{A Family Matter 00} \\
\textbf{In Objective(s): } Tucker Needham ran away from Stellar Bay a few weeks ago to join the Iconoclasts in Amber Heights. His mother Agnes is willing to pay handsomely if you can locate her son and convince him to return. You can begin this quest by talking to Agnes Needham in Stellar Bay, Monarch. Agnes is by the town's south-east exit, visibly shaken and calling for help. Hear her out and offer to find her son to being the quest.\\
\textbf{Out Objective(s): } Amber Heights is the settlement that houses the Iconoclasts on Monarch. If Tucker Needham survived his travels, his mother thinks he'll be there. \\
\textbf{Game Lore: } \\
Agnes Needham\\
\textit{[0] Agnes Needham is a resident of Stellar Bay and the mother of Tucker Needham. [1] Agnes' overprotective style of mothering has led her son, Tucker Needham, to run away from home so he can experience life. [2] Despite Tucker being 42 years old, she still thinks of him as her 'little boy'. [3] You can find her by Stellar Bay's south-east exit, visibly shaken and calling for help.}\\
Iconoclasts\\
\textit{[0] The Iconoclasts are a group of survivalists living in the ruins of Amber Heights on Monarch. [1] They hope to one day tear down the corporate establishment that they believe has brought the colony to the brink of death. [2] The Iconoclasts are a group of idealistic revolutionaries that seek to overthrow the corporate establishment that runs the Halcyon Colony. [3] Based in the ruins of the Amber Heights settlement on Monarch, [4] they are a tenacious group, [5] and share some democratic ideals with Monarch Stellar Industries (MSI) against the more repressive actions of the Board. [6] However, the Iconoclast's anti-corporate nature has put them at odds with MSI, a dispute that threatens to spill into all-out warfare. [7] Given that the Iconoclasts are mostly followers of the Philosophist faith, they have been blacklisted and demonized by the Board as dissenters and anarchists. [8] The group is led by Graham Bryant, a staunch Philosophist. [9] Zora Blackwood, the Iconoclasts' chief of medicine, is also considered a de facto leader of the group, [10] as she was alongside Graham when he founded the Iconoclasts, [11] and almost every member of the Iconoclasts owes her their life in some way. [12] The Iconoclasts maintain a tense relationship with MSI. [13] Despite sharing democratic values and a common desire towards egalitarianism for the people of Monarch and the wider Halcyon colony, [14] MSI's "egalitarian corporate structure" has proven to be at odds with some of the Iconoclasts' more radical, anti-capitalist views. [15] Depending on the actions of the Stranger, this tense relationship can either be resolved, [16] or can spill into a drawn-out and bloody war. [17] The Stranger meets the Iconoclasts in Amber Heights just as the tension between them and MSI is reaching boiling point. [18] They can either side with the Iconoclasts and assist them in storming and taking over Stellar Bay, [19] "solve" the Iconoclast problem for Stellar Bay, [20] or broker peace between the two factions. [21] The Stranger can also have an impact on the leadership of the Iconoclasts - siding with either Graham Bryant or Zora Blackwood. [22] To supplant Graham with Zora, evidence of Graham's involvement in the Amber Heights massacre must be found and presented to Zora. [23] The Van Noys are the Iconoclasts' best unit.}\\
Mantisaur\\
\textit{[0] Mantisaurs are insectoid creatures native to Monarch. [1] They are aggressive, territorial, and very strong. [2] It is possible to deal with them one on one, but it is best to avoid groups of them for your safety. [3] The mantiqueen is the largest breed of Mantisaur.}\\
Monarch\\
\textit{[0] Monarch, previously known as Terra 1, is one of the many moons of the gas giant Olympus and the site of a failed colony. [1] Terra 1 was initially designated as the primary colonization target of the Halcyon system. [2] The Halcyon Holdings Corporate Board had intended to completely terraform the moon, [3] wiping out the local fauna and flora and replacing it with plants and wildlife native to Earth. [4] However, the terraforming process unexpectedly caused the native species to mutate and grow to significantly larger sizes, [5] rendering them more dangerous and severely crippling the colonization effort. [6] Due to the hostile environment which they had created, [7] the Board was forced to enact a Hazard Clause covering the entirety of Terra 1. [8] Public notice of the clause's issuance was sent to everyone operating on Terra 1 and led to the evacuation of almost all corporations from the moon. [9] However, one corporation took advantage of the chaos of the evacuation to exploit a legal loophole which allowed them to, [10] as the last corporation remaining on the planet, [11] acquire the planet from the Board. [12] This corporation, under the leadership of Sanjar Nandi and Graham Bryant subsequently rebranded itself to Monarch Stellar Industries (MSI), [13] in line with the renaming of the planet to 'Monarch'. [14] The actions of MSI earned them the ire of the Board, [15] who retaliated by effectively placing the moon under indefinite embargo, [16] refusing to allow legal transit either in or out. [17] the Board aggressively spread propaganda about Monarch to convince the rest of the population that it was both uninhabited and uninhabitable. [18] This has greatly hampered MSI's attempts to be recognized as a legitimate corporation and is a thorn in the side of its CEO, Sanjar Nandi. [19] Monarch also has an ocean which goes around the moon at the "twilight band". [20] It is where the colonists and Monarch Stellar Industries farm their saltuna.}\\
Raptidon\\
\textit{[0] Raptidons are giant cat/reptile-like creatures that inhabit various planets in Halcyon. [1] They are creatures native to Monarch. [2] however some corporations have illegally imported them to other planets, [3] such as Auntie Cleo who relocated a group of them to Roseway. [4] Raptidons are of corporate interest due to their potential for producing new chemical by-products which, [5] when refined, can be used to create new board approved products.}\\
Stellar Bay\\
\textit{[0] Outside the city walls, the lands were overrun by the native wildlife, as well as marauders and outlaws. [1] Stellar Bay is a company town located on the planet Monarch. It is owned and operated by Monarch Stellar Industries. [2] Stellar Bay is the largest saltuna producer on the Halcyon colony and used to be one of the most important suppliers of this resource.}\\
Sulfur Pits\\
\textit{[0] The Sulfur Pits are a point of interest on the western side of Monarch. [1] They are located southwest of Terra One Publications and directly northeast of the Gunship Crash Site. [2] The Sulfur Pits have a large variety of Raptidons and many deceased marauders. [3] The area consists largely of Sulfur Pits. [4] When an entity comes in contact with a sulfur pit, [5] they receive the acid effect for the duration of touching the pit.}\\
Tucker Needham\\
\textit{[0] Tucker Needham is a former resident of Stellar Bay who left to join the Iconoclasts. [1] Before the quest A Family Matter, he can be found in Amber Heights. [2] Tucker was coddled by his mother from a very young age, [3] the latter insisting that danger lurked around every corner on Monarch. [4] His mother's overprotectiveness extended well into Tucker's adulthood, [5] leading him to seek to be free in any way possible. [6] After hearing Graham Bryant's broadcasts, Tucker left Stellar Bay to be truly free by joining the Iconoclasts at Amber Heights. [7] He is dazzled by Graham's preachings on true unfettered freedom from the corporate way of life and attributes his enthusiasm to his 'childhood trauma'. [8] He is willing to do anything to remain free, even faking his own death to prevent his mother from continuing to send people to look for him.}\\ \bottomrule
\end{tabular}
    \caption{Dialogue from motivating example in \autoref{fig:overview} with all input constraining passages. Full dialogue tree can be found on the next page.}
    \label{fig:family-matter-input}
    
\end{figure*}


\begin{figure*}
     \centering
    \includegraphics[width=\textwidth]{images/dialogs/the_chimerists_last_experiment_01.pdf}
    \caption{Example of longer  dialogue tree in \name{}, containing numerous decision points, cycles and re-entrances.}
    \label{fig:tree-3}
\end{figure*}

\section{Comparison with Other Datasets}
\label{app:comparison}
\citet{DBLP:conf/aiide/StegerenT20, DBLP:conf/fdg/StegerenM21} consider datasets of publicly-available side quest data from RPGs such as \textit{World of Warcraft}.  However, their datasets vary in dialogue and quest coverage; for \textit{WoW} their input is just a quest name and objective, and the generation target is a single-turn, few-sentence quest description spoken by an NPC. heir collect data for the game \textit{TorchLight II} contains quest datapoints with a limited number dialog utterances per quest with no multi-turn interactions or trees.\footnote{\autoref{tab:comparison} describes statistics for the 82 TorchLight II quests that contain both objective annotations and dialogue lines. }
Others of their collected datasets contain complex branching trees but without constraining knowledge.
 The dialogues of LIGHT~\cite{DBLP:conf/emnlp/UrbanekFKJHDRKS19} are more akin to NPC dialogues, though they comprise few-turn linear chains between two characters in self-contained episodes rather than quest-grounded interactions between a player and an NPC serving multiple game purposes. The size of constraining passages on the LIGHT dialogues are also a scale smaller than those of \name{}.
The biographical constraints of \name{} are most similar to that of \textsc{TVStoryGen}~\cite{chen-gimpel-2022-leveraging}, who also pull articles from fandom wiki pages. However, theirs is a story generation dataset where the target is a longform article describing a TV episode.
\section{Example Prompt Constructions for DialogueWriter Model}
\label{sec:app-prompts}
\autoref{fig:method} depicts an overview of our tree linearization and prompt construction method.
\begin{figure*}
\centering
\begin{lrbox}{\myv}\begin{minipage}{\textwidth}
\begin{lstlisting}[basicstyle=\fontsize{6}{7}\selectfont\ttfamily]
source:
the Board's authoritarianism.</s> Sanjar Nandi</s> Sanjar Nandi is the current CEO of Monarch Stellar Industries, based in Stellar Bay.</s> Sanjar began working for MSI at a young age and it was there where he met Graham Bryant, who would eventually become his best friend.</s> Sanjar was ambitious but his attention to detail at the expense of big-picture thinking hampered his efforts within MSI.</s> This led to negative performance reviews regarding his tendency to pad reports and talks with numbers and data,</s> feedback which continues to haunt him many years on.</s> However, the negative feedback did not dampen Sanjar's desire to move up within the company,</s> even donating a kidney to one of the executives in hopes of promotion.</s> Despite his poor performance, Sanjar always showed himself to be a loyal employee of the company.</s> Despite Sanjar's best efforts, he has found it extremely challenging to continue operating MSI on Monarch without the backing of the Board.</s> In order to improve the lives of the people he is responsible for, Sanjar has a plan to rejoin the Board through the use of a BOLT-52 form and proof of another corporate presence on Monarch.</s> He is simultaneously working on a plan to reorganize the Board,</s> hoping that his plans are not found out until MSI has been reinstated.</s> Sanjar choosing to take over as head of MSI rather than dismantling it entirely caused a rift between him and Graham Bryant.</s> The latter started the Iconoclasts,</s> a group dedicated to spreading the word of Philosophism throughout the galaxy,</s> and Sanjar was left in Stellar Bay to run the company and look after the employees who were left behind.</s> He can also tell you more about the planet, that used to be called Terra 1</s> and the reform that he and Monarch Stellar Industries tried to achieve to give more humane working conditions for everyone within.</s> Celia Robbins</s> Celia Robbins is a middle manager for Monarch Stellar Industries and works with Sanjar Nandi at MSI Headquarters in Stellar Bay.</s> Celia has a crush on Sebastian Adams and will buy whatever he has in stock,</s> just as an excuse to talk to him.</s> Unfortunately her apartment is filling up with exotic creature parts and her neighbors are starting to complain about the smell.</s> She is not concerned that she and Sebastian may not have much to talk about,</s> as everyone else in Stellar Bay either smells like saltuna or are her boss.</s> The Stranger can offer to set her and Sebastian up on a date.</s> DIALOG CONTEXT: Sanjar believes another company may be operating on Monarch illegally.</s> If he can get proof, then he could use that as leverage to get MSI readmitted to the Halcyon Board.</s> Talk to Sanjar after completing BOLT with His Name.</s> He reveals the plan is to blackmail The Board into letting them back to the table.</s> Sanjar will reveal he believes another corporation is operating illegally within Monarch, granting you the quest Errors Unseen.</s> He will tell you Catherine is likely supplying them from Fallbrook.</s> He wants you to infiltrate the secret facility and bring back evidence - be it an item or staff. KNOW BY THE END OF THE DIALOG: Sanjar believes Catherine would know where this other corporation is operating.</s> See if you can get the location from her.</s> DIALOG PARTICIPANTS:</s> Sanjar Nandi</s> Celia Robbins</s> Player</s> HISTORY:> Player: <unk>Give him the BOLT-52.> I found the cartridge and deleted that data for you. > Sanjar Nandi: Oh, yes. I'm going to be up all night with this. All those blanks waiting to be filled, boxes waiting to be ticked... > Celia: Try to control yourself, sir. > Sanjar Nandi: Have you any idea how powerful this is? Corporations have been toppled with less. > Player: How exactly is a data cartridge going to help? > Sanjar Nandi: What a question! Bureaucratic micromanagement is the only way anything gets done on Halcyon, and proper documentation is a key part of that. > Sanjar Nandi: For our part, a Bill of Liquidation/Transfer Form-52 will protect our holdings on Monarch by temporarily assigning them to a pass-through entity once we drop our bomb on the Board.</s>
target:
The Board fact: The Board maintains a very tense relationship with MSI, owing to MSI's democratic ideals and their declared ownership of Monarch., The Board fact: Depending on the actions of the Stranger, MSI may be compelled to rebel against the Board's authoritarianism. > Player: Yes! Finally, the Board will get their comeuppance!</s>

---

source:
mire to help her with a comms issue,</s> and inform the player that they are now authorized to trade with Doctor Mfuru.</s> Junlei will appear on the Unreliable if the player helps to set up a date between her and Parvati.</s> When interacted with, she thanks them for welcoming her aboard the ship.</s> Mardets</s> The Mardets or Groundbreaker Security are the security force on the Groundbreaker.</s> The Mardets started as the original security force on Groundbreaker before the Crossing.</s> The original force was made up of a marine detachment from the 77th Marine Expeditionary Unit, Trailward Fleet.</s> Over time people started to call them Mardets because it was easier to say and the name stuck.</s> The Mardets that currently protect Groundbreaker are descended from the original marine detachment and still wear colony ship guard armor as a uniform.</s> Sublight</s> The SubLight Salvage and Shipping Corporation, also known as SubLight Salvage and Shipping or simply SubLight, is a network of "salvagers" with business ties to "transportation" and "waste disposal".</s> A tangled web of contractors and secretive vice presidents make up their official hierarchy,</s> leaving no one to speak on the record about SubLight's more legally dubious activities.</s> SubLight as a business operates activities that would be considered illicit by some,</s> amoral by others,</s> and at odds with the Board,</s> such as the takeover of abandoned space habitats,</s> smuggling contraband,</s> and serving as a front for companies to bypass the infamous Hazard Clause applied on Monarch.</s> According to director Lilya Hagen,</s> operating off of Groundbreaker,</s> SubLight "occupies a legal blind spot.</s> No one knows what [they're] licensed to do,</s> and that gives [their] little business some freedom."</s> Although the Board tolerates and sees the value of SubLight's activities,</s> there are still limits to what the latter can get away with.</s> SubLight maintains an office aboard the Groundbreaker as well as a settlement on Monarch known as Fallbrook.</s> Gunnar Macredd</s> Captain Gunnar MacRedd is a known criminal and leader of a group of outlaws found in the Back Bays of Groundbreaker.</s> MacRedd is a pyromaniac with a criminal record which includes "several counts of flying under the influence,</s> carrying open alcoholic containers,</s> failure to pay docking fees,</s> resisting arrest,</s> and assaulting not one but two officers."</s> He has a lighter which he claims was given to him by Commandant Sanita a few years back when the two had a "carnal understanding".</s> However, Sanita claims that this understanding only consisted of her "kick[ing] his ass from one end of the Groundbreaker to the other".</s> DIALOG CONTEXT: Based on the bounty listing, Captain McRedd was last sighted in the Back Bays.</s> Head there and take him out.</s> You can find Captain MacRedd in the Back Bays area of the Groundbreaker.</s> To get there head down the elevator in the promenade,</s> and you can't miss him.</s> You can pass a Persuade (40) check to get him to put his gun down,</s> otherwise you'll have to kill him and all his guards.</s> If you kill him he drops the Unique Weapon: Montag.</s> You'll get 6000xp and MacRedd's Lighter.</s> If you persuaded him, use Perception to note it says "Sanita" on the lighter.</s> MacRedd will mention it was given to him by Sanita in remembrance of a 'carnal understanding' they had a few years back. KNOW BY THE END OF THE DIALOG: McRedd gave you his lucky lighter to give to Sanita.</s> Go turn it in to resolve his bounty.</s> DIALOG PARTICIPANTS:</s> Gunnar Macredd</s> Player</s> HISTORY:> Captainmacredd: Look at this ripe piece of meat just sizzling on the grill. Yum yum. > Captainmacredd: Time to feed the flames. It's nothing personal. Promise. > Player: You're only getting out of this alive if you play nice.</s>
target:
> Captainmacredd: Look at the courage on this one! Lay it on me, boss. Give me a taste of that fiery, primal spirit.</s>
\end{lstlisting}
\end{minipage}\end{lrbox}
\resizebox{0.95\textwidth}{!}{\usebox\myv}
\caption{Example source and target items used to train and evaluate T5-based SL DialogueWriters. The first item exhibits support facts prepended to the target for the SL Knowledge Selection model.}
\label{fig:t5-prompt}
\end{figure*}

\begin{figure*}
   \centering
\begin{lrbox}{\myv}\begin{minipage}{\textwidth}
\begin{lstlisting}[basicstyle=\fontsize{5}{6}\selectfont\ttfamily]
FACTS:
Iconoclasts
   The Iconoclasts are a group of survivalists living in the ruins of Amber Heights on Monarch.
   They hope to one day tear down the corporate establishment that they believe has brought the colony to the brink of death.
   The Iconoclasts are a group of idealistic revolutionaries that seek to overthrow the corporate establishment that runs the Halcyon Colony.
   Based in the ruins of the Amber Heights settlement on Monarch,
   they are a tenacious group,
   and share some democratic ideals with Monarch Stellar Industries (MSI) against the more repressive actions of the Board.
   However, the Iconoclast's anti-corporate nature has put them at odds with MSI, a dispute that threatens to spill into all-out warfare.
   Given that the Iconoclasts are mostly followers of the Philosophist faith, they have been blacklisted and demonized by the Board as dissenters and anarchists.
   The group is led by Graham Bryant, a staunch Philosophist.
   Zora Blackwood, the Iconoclasts' chief of medicine, is also considered a de facto leader of the group,
   as she was alongside Graham when he founded the Iconoclasts,
   and almost every member of the Iconoclasts owes her their life in some way.
   The Iconoclasts maintain a tense relationship with MSI.
   Despite sharing democratic values and a common desire towards egalitarianism for the people of Monarch and the wider Halcyon colony,
   MSI's "egalitarian corporate structure" has proven to be at odds with some of the Iconoclasts' more radical, anti-capitalist views.
   Depending on the actions of the Stranger, this tense relationship can either be resolved,
   or can spill into a drawn-out and bloody war.
   The Stranger meets the Iconoclasts in Amber Heights just as the tension between them and MSI is reaching boiling point.
   They can either side with the Iconoclasts and assist them in storming and taking over Stellar Bay,
   "solve" the Iconoclast problem for Stellar Bay,
   or broker peace between the two factions.
   The Stranger can also have an impact on the leadership of the Iconoclasts - siding with either Graham Bryant or Zora Blackwood.
   To supplant Graham with Zora, evidence of Graham's involvement in the Amber Heights massacre must be found and presented to Zora.
   The Van Noys are the Iconoclasts' best unit.
Monarch
   Monarch, previously known as Terra 1, is one of the many moons of the gas giant Olympus and the site of a failed colony.
   Terra 1 was initially designated as the primary colonization target of the Halcyon system.
   The Halcyon Holdings Corporate Board had intended to completely terraform the moon,
   wiping out the local fauna and flora and replacing it with plants and wildlife native to Earth.
   However, the terraforming process unexpectedly caused the native species to mutate and grow to significantly larger sizes,
   rendering them more dangerous and severely crippling the colonization effort.
   Due to the hostile environment which they had created,
   the Board was forced to enact a Hazard Clause covering the entirety of Terra 1.
   Public notice of the clause's issuance was sent to everyone operating on Terra 1 and led to the evacuation of almost all corporations from the moon.
   However, one corporation took advantage of the chaos of the evacuation to exploit a legal loophole which allowed them to,
   as the last corporation remaining on the planet,
   acquire the planet from the Board.
   This corporation, under the leadership of Sanjar Nandi and Graham Bryant subsequently rebranded itself to Monarch Stellar Industries (MSI),
   in line with the renaming of the planet to 'Monarch'.
   The actions of MSI earned them the ire of the Board,
   who retaliated by effectively placing the moon under indefinite embargo,
   refusing to allow legal transit either in or out.
   the Board aggressively spread propaganda about Monarch to convince the rest of the population that it was both uninhabited and uninhabitable.
   This has greatly hampered MSI's attempts to be recognized as a legitimate corporation and is a thorn in the side of its CEO, Sanjar Nandi.
   Monarch also has an ocean which goes around the moon at the "twilight band".
   It is where the colonists and Monarch Stellar Industries farm their saltuna.
Raptidon
   Raptidons are giant cat/reptile-like creatures that inhabit various planets in Halcyon.
   They are creatures native to Monarch.
   however some corporations have illegally imported them to other planets,
   such as Auntie Cleo who relocated a group of them to Roseway.
   Raptidons are of corporate interest due to their potential for producing new chemical by-products which,
   when refined, can be used to create new board approved products.
Stellar Bay
   Outside the city walls, the lands were overrun by the native wildlife, as well as marauders and outlaws.
   Stellar Bay is a company town located on the planet Monarch. It is owned and operated by Monarch Stellar Industries.
   Stellar Bay is the largest saltuna producer on the Halcyon colony and used to be one of the most important suppliers of this resource.
Sulfur Pits
   The Sulfur Pits are a point of interest on the western side of Monarch.
   They are located southwest of Terra One Publications and directly northeast of the Gunship Crash Site.
   The Sulfur Pits have a large variety of Raptidons and many deceased marauders.
   The area consists largely of Sulfur Pits.
   When an entity comes in contact with a sulfur pit,
   they receive the acid effect for the duration of touching the pit.
Tucker Needham
   Tucker Needham is a former resident of Stellar Bay who left to join the Iconoclasts.
   Before the quest A Family Matter, he can be found in Amber Heights.
   Tucker was coddled by his mother from a very young age,
   the latter insisting that danger lurked around every corner on Monarch.
   His mother's overprotectiveness extended well into Tucker's adulthood,
   leading him to seek to be free in any way possible.
   After hearing Graham Bryant's broadcasts, Tucker left Stellar Bay to be truly free by joining the Iconoclasts at Amber Heights.
   He is dazzled by Graham's preachings on true unfettered freedom from the corporate way of life and attributes his enthusiasm to his 'childhood trauma'.
   He is willing to do anything to remain free, even faking his own death to prevent his mother from continuing to send people to look for him.
Agnes Needham
   Agnes Needham is a resident of Stellar Bay and the mother of Tucker Needham.
   Agnes' overprotective style of mothering has led her son, Tucker Needham, to run away from home so he can experience life.
   Despite Tucker being 42 years old, she still thinks of him as her 'little boy'.
   You can find her by Stellar Bay's south-east exit, visibly shaken and calling for help.

DIALOG CONTEXT:
Tucker Needham ran away from Stellar Bay a few weeks ago to join the Iconoclasts in Amber Heights.
His mother Agnes is willing to pay handsomely if you can locate her son and convince him to return
   You can begin this quest by talking to Agnes Needham in Stellar Bay, Monarch.
   Agnes is by the town's south-east exit, visibly shaken and calling for help. Hear her out and offer to find her son to being the quest.

KNOW BY THE END OF THE DIALOG:
Amber Heights is the settlement that houses the Iconoclasts on Monarch.
If Tucker Needham survived his travels, his mother thinks he'll be there.

DIALOG PARTICIPANTS:
Agnes Needham, Player

DIALOG:
> Agnes Needham: Oh, thank you for stopping! Everyone acts like nothing's wrong. Like my little boy isn't at risk of being eaten by some vile creature!
> Agnes Needham: Please, you have to help me get my little Tucker back! He ran away and is going to get himself killed! Oh, I just know a raptidon is melting him with acid as we speak!
> Player: Calm down. Take a deep breath, then tell me what happened.
> Agnes Needham: Don't you tell me to calm down! I promised my boy I'd protect him for always. But how can I keep him safe if he's run away?
> Agnes Needham: He ran out into the wilderness a few days ago. I warned him about the raptidons, mantisaurs, and marauders - the toxic sulfur pools, and poisonous plants. But he didn't listen!
> Agnes Needham: Please! Won't you go and find my boy?
> Player: Did you say there was a reward involved with this request?
> Agnes Needham: Well, I... I guess I can't ask you to leave the town walls for free. It is deathly dangerous out there.
> Agnes Needham: I've got some bits saved up for a rainy day. I'll give you every last one if you just bring my Tucker back to me. I won't even be mad at him running off. You tell him, I won't be mad.
> Player: Why would Tucker run away?
> Agnes Needham: He's been pining for an adventure, says he's tired of living cooped up behind the walls. But he doesn't understand how dangerous it is out there.
> Agnes Needham: I warned him. A raptidon would snap him up first chance it got. I just know one's ripped his arm off and is gnawing on his sweet little fingers!
> Agnes Needham: He should've listened to his mama. I promised I'd keep him safe here with me.
> Player: I'll help you find your son.
\end{lstlisting}
\end{minipage}\end{lrbox}
\resizebox{0.95\textwidth}{!}{\usebox\myv}
\caption{Example In-Context Learning (ICL) prompt for GPT-3 based DialogueWriter}
\label{fig:gpt3-prompt}
\end{figure*}

\begin{figure*}
   \centering
\begin{lrbox}{\myv}\begin{minipage}{\textwidth}
\begin{lstlisting}[basicstyle=\fontsize{5}{6}\selectfont\ttfamily]
DIALOG:
Agnes Needham fact: Agnes Needham is a resident of Stellar Bay and the mother of Tucker Needham.
Agnes Needham fact: Agnes' overprotective style of mothering has led her son, Tucker Needham, to run away from home so he can experience life.
Agnes Needham fact: Despite Tucker being 42 years old, she still thinks of him as her 'little boy'.
Agnes Needham fact: You can find her by Stellar Bay's south-east exit, visibly shaken and calling for help.
Tucker Needham fact: the latter insisting that danger lurked around every corner on Monarch.
utterance: > Agnes Needham: Oh, thank you for stopping! Everyone acts like nothing's wrong. Like my little boy isn't at risk of being eaten by some vile creature!

Agnes Needham fact: Agnes Needham is a resident of Stellar Bay and the mother of Tucker Needham.
Agnes Needham fact: Agnes' overprotective style of mothering has led her son, Tucker Needham, to run away from home so he can experience life.
Agnes Needham fact: Despite Tucker being 42 years old, she still thinks of him as her 'little boy'.
Tucker Needham fact: the latter insisting that danger lurked around every corner on Monarch.
Raptidon fact: Raptidons are giant cat/reptile-like creatures that inhabit various planets in Halcyon.
utterance: > Agnes Needham: Please, you have to help me get my little Tucker back! He ran away and is going to get himself killed! Oh, I just know a raptidon is melting him with acid as we speak!

utterance: > Player: Calm down. Take a deep breath, then tell me what happened.

Agnes Needham fact: Agnes' overprotective style of mothering has led her son, Tucker Needham, to run away from home so he can experience life.
utterance: > Agnes Needham: Don't you tell me to calm down! I promised my boy I'd protect him for always. But how can I keep him safe if he's run away?

Agnes Needham fact: Agnes' overprotective style of mothering has led her son, Tucker Needham, to run away from home so he can experience life.
Stellar Bay fact: Outside the city walls, the lands were overrun by the native wildlife, as well as marauders and outlaws.
Raptidon fact: Raptidons are giant cat/reptile-like creatures that inhabit various planets in Halcyon.
Monarch fact: However, the terraforming process unexpectedly caused the native species to mutate and grow to significantly larger sizes,
Monarch fact: rendering them more dangerous and severely crippling the colonization effort.
Mantisaur fact: Mantisaurs are insectoid creatures native to Monarch.
Mantisaur fact: They are aggressive, territorial, and very strong.
Sulfur Pits fact: The Sulfur Pits are a point of interest on the western side of Monarch.
utterance: > Agnes Needham: He ran out into the wilderness a few days ago. I warned him about the raptidons, mantisaurs, and marauders - the toxic sulfur pools, and poisonous plants. But he didn't listen!

Agnes Needham fact: Despite Tucker being 42 years old, she still thinks of him as her 'little boy'.
utterance: > Agnes Needham: Please! Won't you go and find my boy?

utterance: > Player: Did you say there was a reward involved with this request?

Stellar Bay fact: Outside the city walls, the lands were overrun by the native wildlife, as well as marauders and outlaws.
utterance: > Agnes Needham: Well, I... I guess I can't ask you to leave the town walls for free. It is deathly dangerous out there.

utterance: > Agnes Needham: I've got some bits saved up for a rainy day. I'll give you every last one if you just bring my Tucker back to me. I won't even be mad at him running off. You tell him, I won't be mad.

utterance: > Player: Why would Tucker run away?

Tucker Needham fact: the latter insisting that danger lurked around every corner on Monarch.
Tucker Needham fact: leading him to seek to be free in any way possible.
Stellar Bay fact: Outside the city walls, the lands were overrun by the native wildlife, as well as marauders and outlaws.
utterance: > Agnes Needham: He's been pining for an adventure, says he's tired of living cooped up behind the walls. But he doesn't understand how dangerous it is out there.

Tucker Needham fact: the latter insisting that danger lurked around every corner on Monarch.
Raptidon fact: Raptidons are giant cat/reptile-like creatures that inhabit various planets in Halcyon.
utterance: > Agnes Needham: I warned him. A raptidon would snap him up first chance it got. I just know one's ripped his arm off and is gnawing on his sweet little fingers!

Agnes Needham fact: Agnes' overprotective style of mothering has led her son, Tucker Needham, to run away from home so he can experience life.
utterance: > Agnes Needham: He should've listened to his mama. I promised I'd keep him safe here with me.

utterance: > Player: I'll help you find your son.

Agnes Needham fact: Agnes' overprotective style of mothering has led her son, Tucker Needham, to run away from home so he can experience life.
utterance: > Agnes Needham: Thank you! Oh, I know he'll be safe now that someone's able to fetch him home.

utterance: > Agnes Needham: You look for him in Amber Heights, you hear? It's down the road southwest of town. I'm sure he made it that far, I just know it.

Agnes Needham fact: Despite Tucker being 42 years old, she still thinks of him as her 'little boy'.
Tucker Needham fact: After hearing Graham Bryant's broadcasts, Tucker left Stellar Bay to be truly free by joining the Iconoclasts at Amber Heights.
Iconoclasts fact: The Iconoclasts are a group of idealistic revolutionaries that seek to overthrow the corporate establishment that runs the Halcyon Colony.
Iconoclasts fact: Based in the ruins of the Amber Heights settlement on Monarch,
utterance: > Agnes Needham: And if you find any of them Iconoclasts indoctrinating my boy, you punch them in the mouths! Tell them what I think of them luring little boys away from their mamas. It's immoral!
\end{lstlisting}
\end{minipage}\end{lrbox}
\resizebox{0.95\textwidth}{!}{\usebox\myv}
\caption{Example In-Context Learning (ICL) prompt with CoT-style support knowledge selection}
\label{fig:cot-prompt}
\end{figure*}
\autoref{fig:t5-prompt} shows example seq2seq items used to train/evaluate the T5-based supervised learning DialogueWriters.
We list the biographies of participants last so as to truncate them from the context only when all other bios have been removed. Else, biographies are listed in random order (fixed at the onset for full dialogue generation).
\autoref{fig:gpt3-prompt} depicts example prompts shown to GPT-3 based in-context-learning DialogueWriters.  
\section{Model Training}
\label{app:training}
To construct training items, we iterate through the nodes of each gold dialogue tree in a canonical
order $[n_1, \dots n_t]$, where $n_1$ is the tree's start node. We create a separate item with each $n_i$ as the generation target. We construct the subtree $S^{(i)}$ comprised of all nodes $[n_1, \dots, n_{i-1}]$ and all edges between them. We then construct the input/output pair $(Q,B,P,S^{(i)}) \rightarrow n_i$.

\paragraph{Supervised Learning}
To train SL DialogueWriter models, 
for every target node in the training quest dialogues, 
 we construct 5 training examples using different random paths to the node. We train the model for 3 epochs using the default arguments from Hugging Face's example summarization model training script.\footnote{\url{https://github.com/huggingface/transformers/blob/main/examples/pytorch/summarization/run_summarization.py}}. T5 models were trained with a batch size of 1 across 8 Quadro RTX 6000 for an average of 5 hours.

\paragraph{In-Context Learning}
Given a test item, we construct a BM25 index over the training dialogues and use it to construct an $n$-shot ICL prompt 
where $n$ depends on the remaining space available in the context window. 
Few-shot examples are linearized dialogues containing the most possible nodes from the gold tree. Contexts are left-truncated and can start with partial examples.
\section{Human Evaluation Directions}
\label{app:human-eval}
Below, we enumerate the instructions shown to annotators during human evaluation:

\textbf{Coherence:} does the utterance follow naturally from the utterances in the history?
    (1) Utterance is nonsensical or ill-formed.
    (2) Utterance is contradictory of previous utterances in the history.
    (4) Utterance naturally responds to the history.

\textbf{Violation:} does the utterance create contradictions with any of the sentences in the ontology or objective blurbs?
    (1) Yes, explicitly contradicts sentences (list the ids).
    (2-3) (gray area).
    (4) No, utterance is consistent with the ontology.

\textbf{Using the Bio Facts:} does the utterance \textit{make use} of the bio sentences in the ontology?
    (1) Utterance is fully generic and/or ignores the ontology completely, could have been generated had the bio facts not been included.
    (2-3) Utterance shows awareness of ontology, albeit unnaturally or inconsistently.
    (4) Utterance naturally incorporates one or multiple pieces of ontology.
    
\textbf{Using the Objectives:} does the utterance progress the dialogue according to the objective sentences in the prompt?
    (1) Utterance ignores objective, could have been generated had the obj facts not been included.
    (2-3) Utterance shows awareness of quest objectives, albeit unnaturally or inconsistently.
    (4) Utterance naturally incorporates one or multiple quest objective statements.

\section{Full Dialogue Evaluation}
\label{app:tree-eval}

\begin{figure*}
    \centering
    \includegraphics[width=\textwidth]{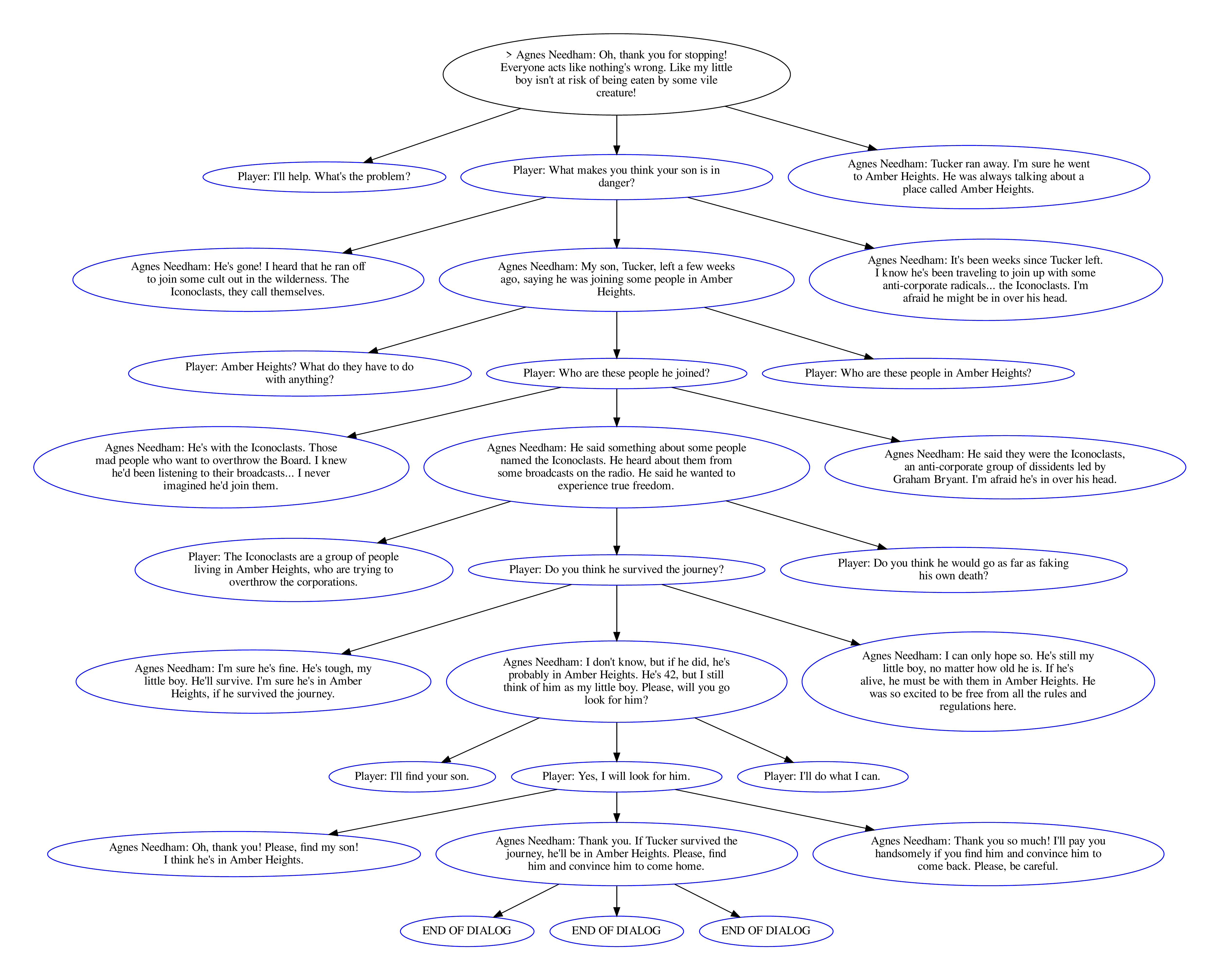}
    \caption{Example dialogue tree generated by the in-context learning knowledge selection DialogueWriter from just the input specifications and starting utterance. Human evaluators were tasked with comparing two such trees and choosing which performed better at a set of qualitative performance criteria. Dialogue follows the specification of the motivating example in \autoref{fig:overview}.}
    \label{fig:tree_eval_afm}
\end{figure*}

\autoref{fig:tree_eval_afm} depicts an example ``spine'' tree shown to evaluators during the end-to-end dialogue evaluation.

The instructions shown to annotators are as follows:

You will replace each `null` value with either "a" or "b", depending on which tree between modela and model b performed better under the following criteria:

\begin{enumerate}
    \item Coherence: do the utterances in the tree create a realistic dialogue between the player character and the NPC?
\item Violations: does the dialogue tree create contradictions with any of the sentences in the ontology or objective blurbs? Does it contradict itself?

\item  Using the Game Lore: does the tree faithfully make of the bio sentences in the ontology, thereby espousing game lore about characters, groups, locations and items?
    
\item  Covering the Objectives: does the dialogue tree play out according to the objective sentences in the prompt?
    
\item Content Suggestion: through generating multiple candidates at each turn, does the dialogue tree effectively propose potential dialogue subtrees that would espouse interesting content?

\item  Engagingness: does the dialogue tree hold your attention and make you want to hear more from the NPC?

\end{enumerate}

\end{document}